\documentclass{article}

\usepackage{microtype}
\usepackage{graphicx}
\usepackage{subfigure}
\usepackage{booktabs} % for professional tables

\usepackage{hyperref}

 \usepackage[accepted]{icml2023}

\usepackage{amsmath}
\usepackage{amssymb}
\usepackage{mathtools}
\usepackage{amsthm}

\usepackage[capitalize,noabbrev]{cleveref}

\theoremstyle{plain}

\theoremstyle{definition}

\theoremstyle{remark}

\usepackage[textsize=tiny]{todonotes}

\icmltitlerunning{Learning Neural PDE Solvers with Parameter-Guided Channel Attention}

\usepackage{comment}
\usepackage{graphicx}
\usepackage{booktabs}       % professional-quality tables
\usepackage{multirow}
\usepackage{xcolor}
\usepackage{algorithm}
\usepackage{algorithmic}
\usepackage{soul}
\usepackage{xspace}
\usepackage{wrapfig}
\usepackage{nicefrac}
\usepackage{tabularx}
\usepackage{bm}
\usepackage{upgreek}

\usepackage[textsize=tiny]{todonotes}
\newcommand{\namednote}[3]{\todo[color=#1!20!white]{\textbf{#2:} #3}}

\newcommand{\francesco}[1]{\namednote{green}{Francesco}{#1}}

\newcommand{\bz}{\bm{z}}
\newcommand{\bx}{\bm{x}}
\newcommand{\by}{\bm{y}}

\newcommand{\bu}{\bm{u}}
\newcommand{\bv}{\bm{v}}

\newcommand{\ba}{\bm{a}}

\newcommand{\btheta}{\bm{\theta}}
\newcommand{\blambda}{\bm{\lambda}}

\newcommand{\bW}{\bm{W}}

\newcommand{\cape}{\textsc{Cape}\@\xspace}
\newcommand{\base}{\textsc{Base}\@\xspace}

\let\orgautoref\autoref

\renewcommand{\autoref}[1]
{%
\def\subsectionautorefname{Sec.}%
\def\sectionautorefname{Sec.}%
\def\equationautorefname{Eq.}%
\def\tableautorefname{Tab.}%
\def\figureautorefname{Fig.}%
\def\subfigureautorefname{Fig.}%
\orgautoref{#1}%
}

\begin{document}

\twocolumn[
%\icmltitle{CAPE: Channel Attention for PDE Parameter Embeddings in \\Neural Surrogate Models}
\icmltitle{Learning Neural PDE Solvers with Parameter-Guided Channel Attention}

% It is OKAY to include author information, even for blind
% submissions: the style file will automatically remove it for you
% unless you've provided the [accepted] option to the icml2023
% package.

% List of affiliations: The first argument should be a (short)
% identifier you will use later to specify author affiliations
% Academic affiliations should list Department, University, City, Region, Country
% Industry affiliations should list Company, City, Region, Country

% You can specify symbols, otherwise they are numbered in order.
% Ideally, you should not use this facility. Affiliations will be numbered
% in order of appearance and this is the preferred way.
\icmlsetsymbol{equal}{*}

\begin{icmlauthorlist}
\icmlauthor{Makoto Takamoto}{nle}
\icmlauthor{Francesco Alesiani}{nle}
\icmlauthor{Mathias Niepert}{nle,stt}
\end{icmlauthorlist}

\icmlaffiliation{nle}{NEC Laboratories Europe, Heidelberg, Germany}
\icmlaffiliation{stt}{University of Stuttgart, Stuttgart, Germany}

\icmlcorrespondingauthor{Makoto Takamoto}{makoto.takamoto@neclab.eu}

% You may provide any keywords that you
% find helpful for describing your paper; these are used to populate
% the "keywords" metadata in the PDF but will not be shown in the document
\icmlkeywords{Machine Learning, ICML}

\vskip 0.3in
]
% this must go after the closing bracket ] following \twocolumn[ ...

% This command actually creates the footnote in the first column
% listing the affiliations and the copyright notice.
% The command takes one argument, which is text to display at the start of the footnote.
% The \icmlEqualContribution command is standard text for equal contribution.
% Remove it (just {}) if you do not need this facility.

\printAffiliationsAndNotice{}  % leave blank if no need to mention equal contribution
%\printAffiliationsAndNotice{\icmlEqualContribution} % otherwise use the standard text.

\begin{abstract}
Scientific Machine Learning (SciML) is concerned with the development of learned emulators of physical systems governed by partial differential equations (PDE). %<-original text
%Accelerating scientific discoveries with machine learning requires learning to emulate physical systems governed by partial differential equations (PDE).
In application domains such as weather forecasting, molecular dynamics, and inverse design, ML-based surrogate models are increasingly used to augment or replace inefficient and often non-differentiable numerical simulation algorithms.
While a number of ML-based methods for approximating the solutions of PDEs have been proposed in recent years, they typically do not adapt to the parameters of the PDEs, making it difficult to generalize to PDE parameters not seen during training. 
We propose a Channel Attention mechanism guided by PDE Parameter Embeddings (\cape) component for neural surrogate models and a simple yet effective curriculum learning strategy. The \cape module can be combined with neural PDE solvers allowing them to adapt to unseen PDE parameters. The curriculum learning strategy provides a seamless transition between teacher-forcing and fully auto-regressive training. 
We compare \cape in conjunction with the curriculum learning strategy using a popular PDE benchmark and obtain consistent and significant improvements over the baseline models. The experiments also show several advantages of \cape, such as its increased ability to generalize to unseen PDE parameters without large increases inference time and parameter count.
An implementation of the method and experiments are available at \url{https://github.com/nec-research/CAPE-ML4Sci}.
\end{abstract}

\section{Introduction}

Many real-world phenomena, ranging from weather forecasts to molecular dynamics and quantum systems, can be modeled with partial differential equations (PDEs). While for some problems the mathematical description of these equations is available, finding its solutions is complex and usually needs some numerical approximations. Numerical simulation methods have been developed for many years and have achieved a high level of accuracy in solving these equations. 
However, numerical methods are resource intensive and time-consuming even when run on larger supercomputers to obtain sufficiently accurate results. Especially high-resolution and high-dimensional hydrodynamic-type field equations are computationally demanding. Even more challenging are simulations with various PDE parameters since a numerical simulation is required for each of the initial conditions and for each of the PDE parameter's configurations. 

Recently, there has been a rapidly growing interest in machine learning methods for the problem of solving PDEs due to their various applications in science and engineering \cite{guo2016convolutional,lusch2018deep,sirignano2018dgm,raissi2018deep,kim2019deep,hsieh2019learning,bar2019learning,bhatnagar2019prediction,pfaff2020learning,wang2020towards,khoo2021solving}. For example, several prior studies reported that ML models can estimate solutions more efficiently than classical numerical simulators \citep{li2020fourier,stachenfeld2021learned}. Moreover, using neural networks as surrogate models allows us to compute derivatives with respect to the input variables. Differentiable surrogate models enable the use of automatic differentiation to solve inverse problems which have numerous real-world applications but are difficult to solve using traditional numerical methods \citep{coros2013computational,2022arXiv220200728A}. A considerable number of papers have shown the advantage of ML-based surrogate models~\citep{li2020neural,li2020fourier,stachenfeld2021learned,lu2021learning}. The majority of these methods, however, are purely data-driven, which does not allow us to change PDE parameters. 
The existing approaches taking PDE parameters into account, are tailored to specific neural network architectures. For instance, message-passing PDE solvers~\cite{brandstetter2022message} use PDE parameters as input but transform these parameters into node embedding and, therefore, cannot be used with other methods, in particular, CNN based models. 
%Although a few models are taking into account PDE parameters, they are tailored to specific neural networks and cannot be used with other state-of-the-art methods. 
%This makes it difficult for the SciML community to develop models with high generalization capability not only for the initial conditions but both for different types of PDEs \emph{and} PDE parameters. 
To overcome the shortcomings of existing data-driven SciML models, a straightforward approach would include the PDE parameters as additional input. 
However, this naive method requires modification of the  \base network which is potentially harmful to its accuracy. An alternative approach attaches an external parameter embedding module to the network. However, there are too many possible module structures and methods to provide the embedded parameter information to the base network, and it is in general non-trivial to select the best one. Still, to ensure a direct comparison, we implemented an extension of said node embedding method to convolution-based methods.

\begin{figure*}[t!]
  \centering
  \includegraphics[width=0.95\textwidth]{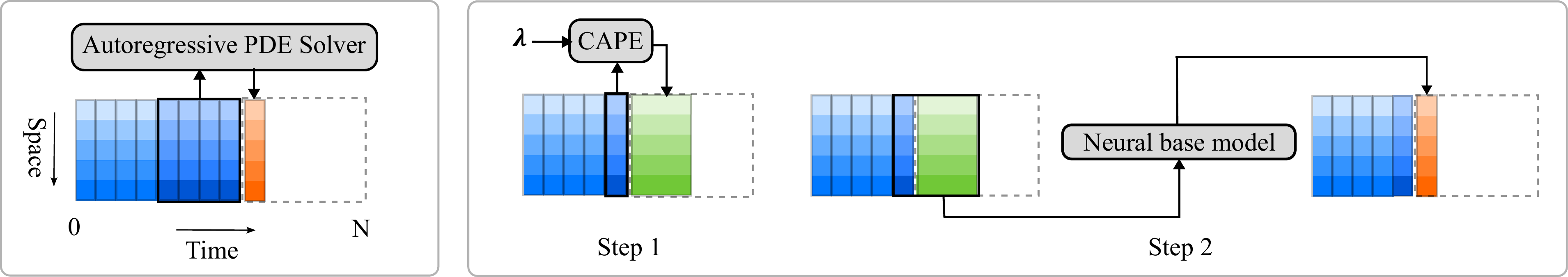}
  \caption{The standard autoregressive approach (left) and the proposed \cape approach (right) which consists of two interdependent steps.}
  \label{fig:Approaches}
\end{figure*}

We propose a new and effective parameter embedding module by utilizing the channel-attention method inspired by classical numerical solvers using the implicit discretization method (see \autoref{overall} and \autoref{sec:cape_structure}).
The crucial idea is that a neural network generates intermediate (approximated) field data for future time steps which are then interpolated by a  \base model such as the FNO \citep{li2020fourier} to predict the field data for the next time step. \cape can be combined with any existing autoregressive neural PDE solvers. \autoref{fig:Approaches} illustrates the proposed \cape framework.

We make the following contributions. First, we propose a \cape module which can be combined with any existing neural PDE solvers and can effectively transfer the PDE parameter information to the base network (\base).
Second, we propose a simple but effective curriculum learning strategy that seamlessly bridges the teacher-forcing and auto-regressive methods.
Third, we perform extensive experiments using various PDEs with a large number of different parameters evaluating the effectiveness and efficiency of the proposed method in comparison with state-of-the-art methods.

\section{\cape: A Framework for Neural PDE Solvers}
\label{sec:method}

\subsection{Background: Partial Differential Equations}
\label{sec:pde_def}

Following the notation by~\cite{brandstetter2022message}, we consider Partial Differential Equations (PDEs) over time dimension $t \in [0,T]$ and over spatial dimensions $\bx=[x_1,\dots,x_D] \in \mathbb{X} \subseteq \mathbb{R}^D$ which can be written as 
\begin{align}
    & \partial_t \bu = F(t,\bx, \bu, \partial_{\bx} \bu, \partial_{\bx,\bx} \bu, \dots ), \\  %~~~ (t,\bx) \in  [0,T] \times \mathbb{X} \\
    & \bu(0,\bx) =  \bu^0(\bx), \  \bx \in \mathbb{X}, ~~~ B[\bu](t,\bx) = 0, %~~~  (t,\bx) \in  [0,T] \times \partial \mathbb{X} 
\end{align}
where $(t,\bx) \in  [0,T] \times \partial \mathbb{X}$, and $\bu : [0, T ] \times \mathbb{X} \to \mathbb{R}^{c}$ is the solution of the PDE, where $c$ is the field dimension, used to describe various field quantities such as velocity, pressure, and density, while $\bu^0(x)$ is the initial condition at time $t = 0$, and $B[\bu](t, \bx) = 0$ are the boundary conditions at $\bx$ in $\partial \mathbb{X}$, which is the boundary of the domain $\mathbb{X}$. Here, $\partial_{\bx} \bu, \partial_{\bx\bx} \bu, \dots$ are the partial derivatives of the solution $\bu$ with respect to the domain, while $\partial_t \bu$ is the partial derivative with respect to time. The functional $F$ describes the possibly non-linear interactions between the PDE's terms.    

\subsection{Problem Definition}
\label{sec:training_def}
We consider PDEs (\autoref{sec:pde_def}) whose solution is described as a temporal sequence of field data $\left\{ \bu^{k} \right\}_{k=0,...,N} := \bu^{0}, \bu^{1}, ..., \bu^{N}$ where $\bu^{k}$ is the field data at time step $t_k$, that is, the state of the physical system governed by the PDE under consideration at time $t_k$ discretized using $\Delta t = T/N$.  Each $\bu \in \mathcal{X} \subseteq \mathbb{R}^{c \times x_1, ..., x_D}$ represents the field tensor data with $c$, the number of physical variables such as density and velocity, and $x_i$ the spatial dimensions of the $i$-th coordinate. For example, for a 1-d problem we have $\mathcal{X} \subseteq \mathbb{R}^{c \times x_1}$, for a 2-d problem $\mathcal{X} \subseteq \mathbb{R}^{c \times x_1 \times x_2}$, and for a 3-d problem $\mathcal{X} \subseteq \mathbb{R}^{c \times  x_1 \times x_2 \times x_3}$.  We will often refer to $c$ as the channel dimension. 
We aim to emulate numerical simulators of PDEs which iteratively  map $\mathcal{M}: \mathcal{X} \to \mathcal{X}$ from $\bu^{k}$ to $\bu^{k+1}$. 
The emulator (or surrogate model) is a learnable function modeled as a neural network $\mathrm{NN}$ with weights $\btheta$. We refer to the parameters of a neural network as weights to avoid a conflict in terminology with the parameters of PDEs.  
In the following, we denote the emulator's prediction at time index $k$ as $\tilde{\bu}^{k}$. Auto-regressive neural networks predict the next time step's field data based on a sequence of field data tensors of length $\ell$
\[\tilde{\bu}^{k+1} = \mathrm{NN}(\tilde{\bu}^{k-\ell+1}, ..., \tilde{\bu}^{k}; \btheta).\]
%\[\tilde{\bu}^{k+1} = \mathrm{NN}(\underbrace{\tilde{\bu}^{k-\ell+1}, ..., \tilde{\bu}^{k}}_{\ell}; \btheta).\]
%\footnote{We always have that at the initial condition $\tilde{\bu}^{(0)} = \bu^{(0)}$}\] 
Given the length of the input sequence $N \in \mathbb{N}$, and an initial input sequence $\left(\bu^{0}, ..., \bu^{\ell-1}\right) = \left(\tilde{\bu}^{0}, ..., \tilde{\bu}^{\ell-1}\right)$ of length $\ell < N$, the ML model auto-regressively generates the remaining sequence $\left(\tilde{\bu}^{\ell}, ..., \tilde{\bu}^{N}\right)$. 
The training loss is typically the normalised root-mean-square error (RMSE) between the predicted and the true field data tensors 
\begin{equation}
    \mathbf{L}(\btheta) = \sum_{k=\ell}^{N} {\rm nRMSE}\left(\bu^{k}, \ \tilde{\bu}^{k}\right) 
    \equiv \sum_{k=\ell}^{N} \frac{||\tilde{\bu}^k - \bu^k||_2}{||\bu^k||_2},
    \label{eq:loss}
\end{equation}
where $||u||_2$ is the $L_2$-norm of a (vector-valued)  variable $u$. 
Since we are training an auto-regressive neural network, the gradients of the above loss can be backpropagated in time in various ways. We discuss this in the following sections. Figure~\ref{fig:Approaches}(left) illustrates this auto-regressive approach to solving PDEs. In the vast majority of experimental setups, the assumption is made that $\ell>1$, 
% that is, 
and, therefore,   
% that a sequence of length greater than 1 is available as input to make the prediction. This implies 
an initial input sequence of length $\ell$ is available to the model; 
% which in practice 
in practice, this 
would require a numerical simulation to be run for $\ell-1$-time steps from the initial condition 
and for each PDE parameter $\blambda$. 
% and for the PDE parameter $\blambda$. 
The main idea of \cape is to learn to generate these sequences based on the current field data and parameter values $\blambda$ and use those as input to an off-the-shelf neural surrogate model such as an FNO \citep{li2020fourier} or an U-Net \citep{RonnebergerUNet2015} to perform a complex interpolation. 

\subsection{Combining Neural PDE Solvers  with the \cape Module}
\label{overall}

The proposed approach is motivated by the need for neural PDE solvers to generalize to PDE parameters unseen during training. 
We propose \cape, a novel neural network architecture that takes the prior state of the system $\tilde{\bu}^{k}$ and PDE parameters $\blambda$ as input and predicts the $\ell$-\emph{intermediate} future states
$\left\{ \hat{\bu}_{\rm cape}^{k \to k + i} \right\}_{i=1, ..., \ell} = \text{\cape} (\bu^k, \blambda; \btheta_\text{\cape})$\footnote{
  In principle it is possible for \cape to also predict field data of past time steps: $\left\{ \hat{\bu}_{\rm cape}^{k \to k + i} \right\}_{i=\pm 1, ..., \pm \ell}$.}. The output of \cape is then used by the \base network. The overall structure is provided in \autoref{fig:Approaches}(right). 
The intuition behind this approach is that the intermediate future states capture information about the PDE parameters' impact by attending to the results of the convolutional operations. While we do not change the architecture of the base neural PDE solvers, we propose to use them to predict, given the past temporal states and the intermediate future states, the state for the next time step. This is contrary to the typical use of neural PDE solvers. The base network is trained jointly with the \cape module. 
As shown in \autoref{sec:experiments}, this choice improves the prediction capability of the \base network. 

During training, 
the output of \cape is augmented with an the additional loss term
% \begin{equation}
%     \label{eq:cape_loss}
%     \mathbf{L}_{\textrm{cape}}(\btheta) = \sum_{k=\ell}^{N} \sum_{i=1}^{{\rm min}(\ell, N-k)} \mathrm{MSE}\left(\bu^{k+i}, \hat{\bu}^{(k),k+i}_{\textrm{cape}}(\bu^{k}, \blambda)\right),
% \end{equation}
%\begin{align}   
%    \label{eq:cape_loss}
    % &\mathbf{L}_{\textrm{cape}}(\btheta) 
 $$\mathbf{L}_{\textrm{cape}}(\btheta_\text{\cape}) 
    = \sum_{k=\ell}^{N} \sum_{i=1}^{{\rm min}(\ell, N-k)} \mathrm{nRMSE}\left(\hat{\bu}_{\rm cape}^{k \to  k+i},{\bu}^{k+i} \right),$$
    % \\ &\left\{ \hat{\bu}_{\rm cape}^{k + i} \right\}_{i=1, ..., \ell}  = \text{\cape} (\bu^k, \blambda; \theta_\text{\cape})
%\end{align}
which forces the \cape module to predict a temporal sequence of future field data $\left\{\bu^{k+i}\right\}_{i=1,...,\ell}$. 

Finally, the intermediate sequence $\left\{ \hat{\bu}_{\rm cape}^{k \to k + i} \right\}_{i=1, ..., \ell}$ is  concatenated with $\bu^k$, the field data at time $t_k$, and given to the base network to make the final prediction. 
In summary, the \cape module transforms the input variables $\left\{ \bu^k, \blambda \right\}$ into temporal-sequential intermediate field data $\{\bu^k, \hat{\bu}^{k \to k + 1}_{\rm cape}, \dots, \hat{\bu}^{k \to k + \ell}_{\rm cape}  \}$ which is then interpolated by the base neural network. Before we introduce the inductive bias of the \cape module, we motivate the general approach from a classical numerical simulation perspective. 

%\textbf{\cape as style interpolation.} 
\textbf{\cape as an implicit discretization method.} 
Let us consider the following simple PDE: 
\begin{equation}
  \partial_t \bu = F(\bu; \blambda)
  \label{eq:pde_ex}
  . 
\end{equation}
The base neural network can be expressed with the equation
% \begin{equation}
%    \bu^{k+1} = f_{\rm base}\left(\bu^{k}, \left\{ \hat{\bu}_{\rm cape}^{(k),k+i} \right\}_{i=1,..,\ell} \right)
%    \label{eq:update}
%    ,
% \end{equation}
\begin{equation}
   \bu^{k+1} = f_{\rm base}\left(\bu^{k}, \left\{ \hat{\bu}_{\rm cape}^{k \to k+i} \right\}_{i=1,..,\ell}; \btheta_\text{\base} \right)
   \label{eq:update}
   ,
\end{equation}
where $f_{\rm base}$ is the function expressed by the base neural surrogate model and 
% $\{ \hat{\bu}_{\rm cape}^{(k),k+i}\}_{i=1,\ell} = f_{\rm cape}(\bu^{k}; \blambda)$ 
% $\{ \hat{\bu}_{\rm cape}^{(k),k+i}\}_{i=1,\ell} = f_{\rm cape}(\bu^{k}; \blambda)$ 
\begin{equation}
\left\{ \hat{\bu}_{\rm cape}^{k \to k + i} \right\}_{i=1, ..., \ell} = \text{\cape} (\bu^k, \blambda; \btheta_\text{\cape})
\end{equation}
are the approximated intermediate future states predicted by the \cape module generated for time index $k$. 
%From the ML perspective, CAPE can be interpreted as a pre-processor network learning behavior in terms of the PDE parameter $\lambda$, and \autoref{eq:update} can be interpreted as a modificator network generating closer style to the present data, $X^t$. On the other hand, those equations can be reinterpreted from a numerical method perspective with which \autoref{eq:pde_ex} can be expressed as: 
The \cape module acts as a pre-processor network providing sequential data $\bu^{k}, \{\bu^{k+i} \}_{i=1,...,\ell}$, while \autoref{eq:update} can be interpreted as an interpolation network whose input is the current state of the physical system $\bu^{k}$ and future extrapolated states of the system $\{\hat{\bu}_{\rm cape}^{k+i} \}_{i=1,...,\ell}$ (see the last figure of \autoref{fig:vis}). 

%\textbf{\cape as implicit discretization method.} 
Now, these equations can be 
% reinterpreted 
understood
from a numerical method perspective, where using  
% with
implicit discretization \citep{anderson2016computational}, \autoref{eq:pde_ex} reduces to: 
\begin{equation}
  \bu^{k+1} = \bu^k + \Delta t F(\bu^{k+1}; \blambda) \equiv \tilde{F}(\bu^k, \bu^{k+1}; \blambda)
  \label{eq:pde_implicit}
  .
\end{equation}
\autoref{eq:update} with $\ell=1$ can be seen as a neural network approximation of the function $\tilde{F}$ used withing the implicit method of \autoref{eq:pde_implicit}. 
When $\ell > 1$ \cape can therefore be seen as a generalized ML-based variant of the implicit method for solving PDEs. 
% Moreover, for $k > 1$, \cape can be interpreted as a generalized ML-based variant of the implicit method. 

%\textcolor{red}{%
Generally, the implicit method is known to be independent of the Courant–Friedrichs–Lewy (CFL) condition \citep{lewy1928partiellen} which enables it to utilize a larger time step size, $\Delta t$, than the explicit method. However, the implicit method usually necessitates either an iterative method or a computationally expensive matrix inversion to obtain the value of $u^{k+1}$ due to the existence of the unknown variable on the right-hand side of the equation. In contrast, our CAPE method allows us to evaluate $u^{k+1}$ in \autoref{eq:pde_implicit} in a data-driven manner. We consider that our approach can potentially result in more accurate and stable predictions of $u^{k+1}$ even using a large time step size by reflecting the implicit method's behavior but with significantly reduced numerical costs.
%}

\begin{figure*}[t!]
  \centering
  \includegraphics[width=0.8\textwidth]{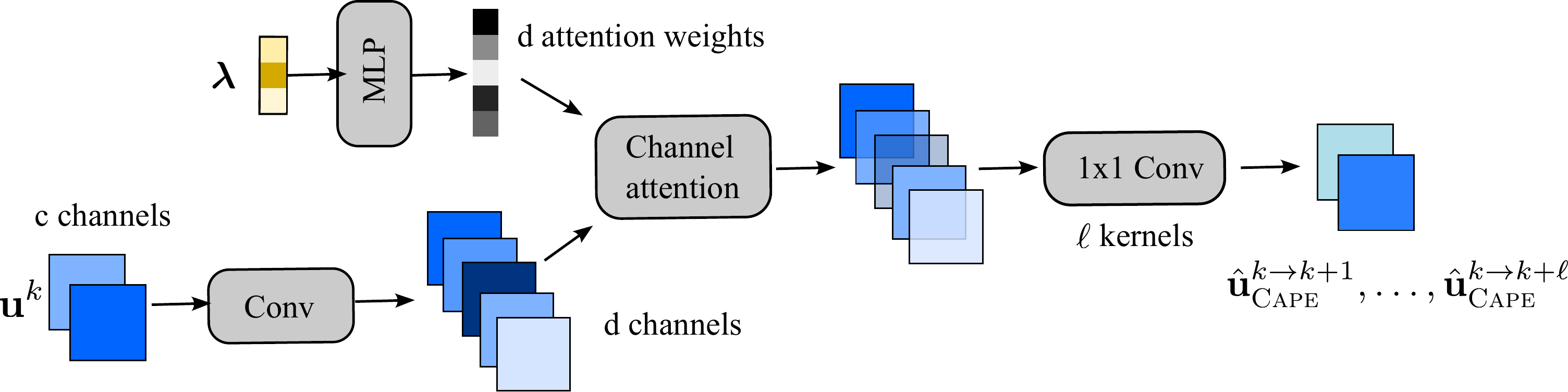}
  \caption{The \cape module for one type of convolution (residual connections are omitted).}
  \label{fig:CAPE}
\end{figure*}

\subsection{PDE Parameter-Guided Channel Attention}
\label{sec:cape_structure}

\cape computes $3$ different $d$-dimensional channel attention masks $\ba_\alpha \in \mathbb{R}^{d}, \alpha=1,2,3$ from the parameters of the PDE $\blambda$ using a $2$-layer MLP 
% $\ba_{\alpha} = \bW_{2, \alpha} \sigma(\bW_{1,\alpha} \blambda),$
\begin{equation}
   \ba_{\alpha} = \bW_{2, \alpha} \sigma(\bW_{1,\alpha} \blambda),
\end{equation}
where 
%\textcolor{red}{%
$\bW_{1/2, \alpha}$ is the weight matrics of the MLPs
%}%
, $d$ is the channel dimension in the feature space and $\sigma$ is the GeLU activation function \citep{hendrycks2016gaussian}. 
%\textcolor{red}{%
Here the bias terms are omitted for simplicity
%}%
. $\bW_\alpha=(\bW_{1,_\alpha},\bW_{2,_\alpha})$ are the weights associated with three operators: a $1\times1$-convolution ($g_1$), a depth-wise convolution ($g_2$), and a spectral convolution \citep{li2020fourier}  ($g_3$), that are used to compute the tensor representations $\bz_\alpha^k \in \mathbb{R}^{d \times n_x \dots}$ as $\bz_\alpha^k = g_\alpha(\bu^k,\bW_\alpha)$.
% \begin{equation}
%     \bz_\alpha^k = g_\alpha(\bu^k,\bW_\alpha)
% \end{equation}
The tensors are then multiplied by the attention
\begin{equation}
    \bv_\alpha^k = \ba_\alpha^k \odot_1 \bz_\alpha^k 
\end{equation}
using the Hadamard operator ($\odot_1$) over the first dimension (the channel dimension) which is equivalent to the broadcast operation of ML programming languages, 
%\textcolor{red}{%
such as Numpy and PyTorch
%}%
. 
The three convolutions can be interpreted as a finite difference method since convolution operations accumulate local information of a mesh, which, in principle, can simulate local interactions such as advection and diffusion. Intuitively, channel attention is equivalent to choosing an appropriate physical process for each PDE parameter. A similar mechanism has been proposed for visual tasks, called the squeeze-and-excitation networks~\citep{hu2018squeeze} which enhances useful channels of the feature vector of convolutional networks through an attention mechanism.
The feature $\bv_\alpha^k \in \mathbb{R}^{d \times n_x \dots}, \alpha=1,,2,3$ are combined to form an intermediate feature $\by^k \in \mathbb{R}^{c \times \ell \times n_x \dots}$ as
% \begin{equation}
%     \by^k = h_{1\times1,b} \left(\sigma\left(h_{1 \times 1,a}(\bu^{k}) + \sum_{\alpha} \bv_\alpha^k \right) \right) \in \mathbb{R}^{c \times \ell \times n_x \dots}
% \end{equation}
\begin{equation}
    \by^k = h_{1\times 1, d \rightarrow c \times \ell} \left(\sigma\left(h_{1 \times 1, c \rightarrow d}(\bu^{k}) + \sum_{\alpha} \bv_\alpha^k \right) \right) 
    \label{eq:cape_conv}
\end{equation}
where $h_{1\times1, *}$ are $1 \times 1$ convolutions that adjust the number of dimensions, in particular $h_{1\times 1, c \rightarrow d}: c \times n_x \dots \to  d \times n_x \dots$, while $h_{1 \times 1, d \rightarrow c \times \ell}: d \times n_x \dots \to  c \times \ell \times n_x \dots$. 
% and $\bu^{}$ is the $n$-th component of the temporal dimension of $y$ after reshaping it into a data tensor of size 
% $b \times c \times \ell \times N_X$
Finally, the sequence of predictions is computed 
\begin{equation}
    \left\{ \bu^{k \to k+i}_{\rm cape} \right\}_{i=1,...,\ell} = (\bu^{k} + {\rm LayerNorm}(\by_i^{k}))_{i=1,...,\ell}
  \label{eq:CAPE_output}
\end{equation}
where $\by_i^{k}$ is the $i$-th element of the data tensor $\by^{k}$, selected from the second dimension 
\footnote{
   We found that in the case of FNO, the LayerNormalization in \autoref{eq:CAPE_output} is harmful (see also \autoref{tab:ablation_CAPE_structure}). 
   We also found that in the case of 2D NS we obtain improved results when modifying the right-hand side of \autoref{eq:CAPE_output} as: $\bu^{k} (1 + {\rm LN}(\by_i^{k}))$ and we applied it to obtain the 2D NS results in this paper.
}. For the sake of presentation, we omitted the batch dimension. 
Figure~\ref{fig:CAPE} illustrates the architecture of the \cape module \footnote{The results with other possible \cape structures are provided in \autoref{sec:ablation_CAPE_structure} which shows our proposed structure is the best choice.}. Visualization of the kernel after channel-attention and a short discussion of the role of the channel-attention in \cape is provided in \autoref{sec:vis-attn}. 

\begin{table*}[t]
\small
\centering
    \begin{tabular}{lll}
        \toprule
        PDE & training parameters  & testing (unseen) parameters in \autoref{fig:Generalization} \\
        \midrule
        1D Advection & $\beta = (0.2, 0.4, 0.7, 2.0, 4.0)$ & $\beta = (0.1, 1.0,7.0)$ \\
        1D Burgers & $\nu = (0.002, 0.007, 0.02, 0.04, 0.2, 0.4, 2.0)$ & $\nu = (0.001, 0.01, 0.1, 1.0, 4.0)$ \\
        2D NS & $\eta = \zeta=(10^{-8}, 0.001, 0.004, 0.01, 0.04, 0.1)$ & $\eta=\zeta=(0.007, 0.07)$ \\        
        \bottomrule
    \end{tabular}
    \caption{PDE parameters used in the experiments.\label{tab:PDE_params}}
    \vspace{-0.2cm}
\end{table*}   

\subsection{Curriculum Learning}
\label{sec:crc}
\begin{figure}[!]
  \centering
  \includegraphics[width=0.3\textwidth]{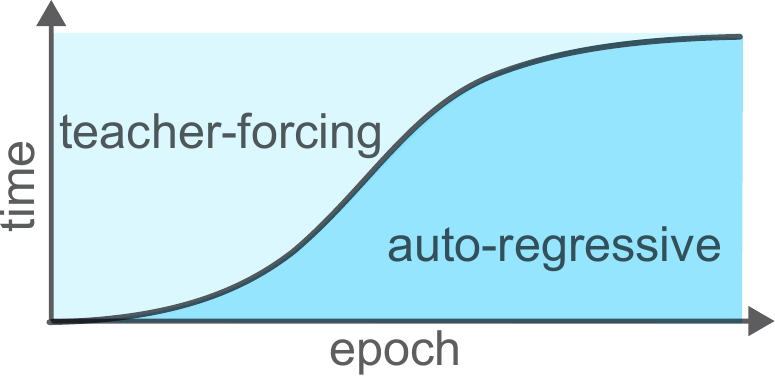}
  \caption{The proposed curriculum learning strategy leads to a smooth transition between one time-step learning ({\it teacher-forcing}) and fully autoregressive training ({\it auto-regressive}).
  % where first the network is trained to predict one time-step ({\it teacher-forcing}), while later longer time span ({\it auto-regressive}).
  }
  \label{fig:CAPE-curriculum-learning}
  % \vspace{-.7cm}
  \vspace{-.5cm}  
\end{figure}
% \begin{comment}
% At training time the model predicts the full temporal sequence $\{ \bu^k \}_{k=0,..,N}$ for one input $\bu^{0}$.
For each initial condition $\bu^{0}$, the \base and \cape models (referred to as $\mathrm{NN}(\bu^k; \btheta)$) jointly predict the full temporal sequence $\{ \bu^{k} \}_{k=1,..,N}$.
We propose the following {\it curriculum learning} strategy, \autoref{fig:CAPE-curriculum-learning}. 
% In the early stage of training, 
For each training epoch, we split the temporal sequence into two parts. For the first part $(\bu^0,...,\bu^{k_{\rm trans}})$ we use {\it auto-regressive} training using $\tilde{\bu}^k$, the prediction of the model at time index $k$, as input to predict the solution for time index $k+1$, that is, $\tilde{\bu}^{k+1} = \mathrm{NN}(\tilde{\bu}^k; \btheta)$.  
For the second part $(\bu^{k_{\rm trans}+1}, ..., \bu^N)$, we train the model using 
{\it teacher-forcing}. The teacher-forcing strategy \citep{williams1989learning,NIPS2015_e995f98d} computes the prediction for time index $k+1$ using a noisy version of the true value at time index $k$, that is, $\tilde{\bu}^{k+1} = \mathrm{NN}(\bu^k + \epsilon; \btheta)$ where $\epsilon$ is random noise increasing the stability at inference time \citep{sanchez2018graph,sanchez2020learning,pfaff2020learning,stachenfeld2021learned}. The  time index $k_{\rm trans}$ determines the time step where we switch from auto-regressive training to teacher-forcing and is computed using the following monotonically increasing function of the epoch number $n$
\begin{align}
    k_{\rm trans} &=  \left\lfloor \frac{N}{2} \left( 1 + \tanh{\left[ \frac{\frac{n}{N} - 0.5}{\Delta}\right]} \right)\right\rfloor, 
    \label{eq:CRC}
\end{align}
where 
% $x(n) = \frac{n}{N}$, 
$N$ is the total epoch number, and $\Delta$ is a hyper-parameter controlling the steepness of the transition function. A plot of the function is provided in \autoref{fig:t-trans}) and 
%, such that at the beginning we favor teacher-forcing, which is more stable, and later in training, we favor auto-regressive, which exposes the model to outputs with accumulated errors
a detailed algorithm of the training strategy is provided in Appendix (Algorithm \autoref{alg:CRC}). 

The strategy is based on the following two assumptions: (1) the prediction error decreases as the number of training epochs increases, (2) the accumulated error increases as the number of auto-regressive rollout steps increases. 
Teacher-forcing training is usually more stable since it avoids the accumulation of prediction errors and should be used exclusively in the first phase of training. The auto-regressive strategy simulates the behavior at test time and exposes the model to inputs that evolved further from the true data, making it more robust to error accumulation. For the same reasons, however, it tends to be less stable, especially in the early phase of training. The proposed curriculum-learning strategy is used to combine the advantages of both approaches.

\section{Experiments}
\label{sec:experiments}
%\subsection{Experiment Setups and Datasets}

We used datasets provided by PDEBench \citep{PDEBenchDataset} a benchmark for SciML from which we selected the following PDEs
\footnote{
An additional experiment results conducted on 2D Burgers equation is provided in \autoref{sec:2d-burgers}. 
}:

\textbf{1D Advection Equation.}
This equation describes the pure-advection of waves 
\begin{align}
     \partial_t u(t,x) + \beta \partial_x  u(t,x) &= 0, 
\end{align}
where $\beta$ is the PDE parameter describing advection velocity. 
The exact solution of this equation is: $u(t, x) = u_0(t, x - \beta t)$ where $u_0$ is the initial condition. Hence, this PDE can be used to check if the ML models understand the property of advection, updating the solution by just advecting the initial profile without changing it. 

\textbf{1D Burgers Equation.}
Burgers' equation is a mathematical model equation simulating the non-linearity and diffusivity in the hydrodynamic equation by a scalar variable 
\begin{align}
    \partial_t u(t,x) + u(t, x) \partial_x u(t,x) &= \nu/\pi \partial_{xx} u(t,x),
    \label{eq:bgs}
\end{align}
where $\nu$ is the diffusion coefficient and the parameter of this equation. This PDE can be used to check if ML models can understand the non-linear behavior from the second term in the left-hand side of \autoref{eq:bgs} and the diffusion process whose strength is controlled by the parameter $\nu$.

\textbf{2D Compressible Navier-Stokes Equations (2D NS).}
The compressible Navier-Stokes equations (NS eqs., in the following) is one of the basic physics equations describing classical fluid dynamics  
 \begin{align}
     \partial_t \rho + \nabla \cdot (\rho \textbf{v}) &= 0, 
     \label{eq:cnast-1} \\
     \rho (\partial_t \textbf{v} + \textbf{v} \cdot \nabla \textbf{v}) &= - \nabla p + \eta \triangle \textbf{v} + (\zeta + \eta/3) \nabla (\nabla \cdot \textbf{v}),
     \label{eq:cnast-2}\\
     \partial_t \left[ \epsilon + \frac{\rho v^2}{2} \right] &+ \nabla \cdot \left[ \left(\epsilon + p + \frac{\rho v^2}{2} \right) \bf{v} - \bf{v} \cdot \sigma' \right] = 0,\label{eq:cnast-3}
 \end{align}
where $\rho$ is the mass density, 
${\bf v}$ is the velocity, 
$p$ is the gas pressure, 
$\epsilon = p/(\Gamma - 1)$ is the internal energy, 
$\Gamma = 5/3$, 
$\sigma'$ is the viscous stress tensor, 
and $\eta, \zeta$ are the shear and bulk viscosity, respectively. 
%Thei is highly non-linear system PDEs, and can be used for a tough test if ML models can understand such a real-world complex PDEs. 

\begin{figure*}[t!]
    \centering
    \includegraphics[width=0.31\textwidth]{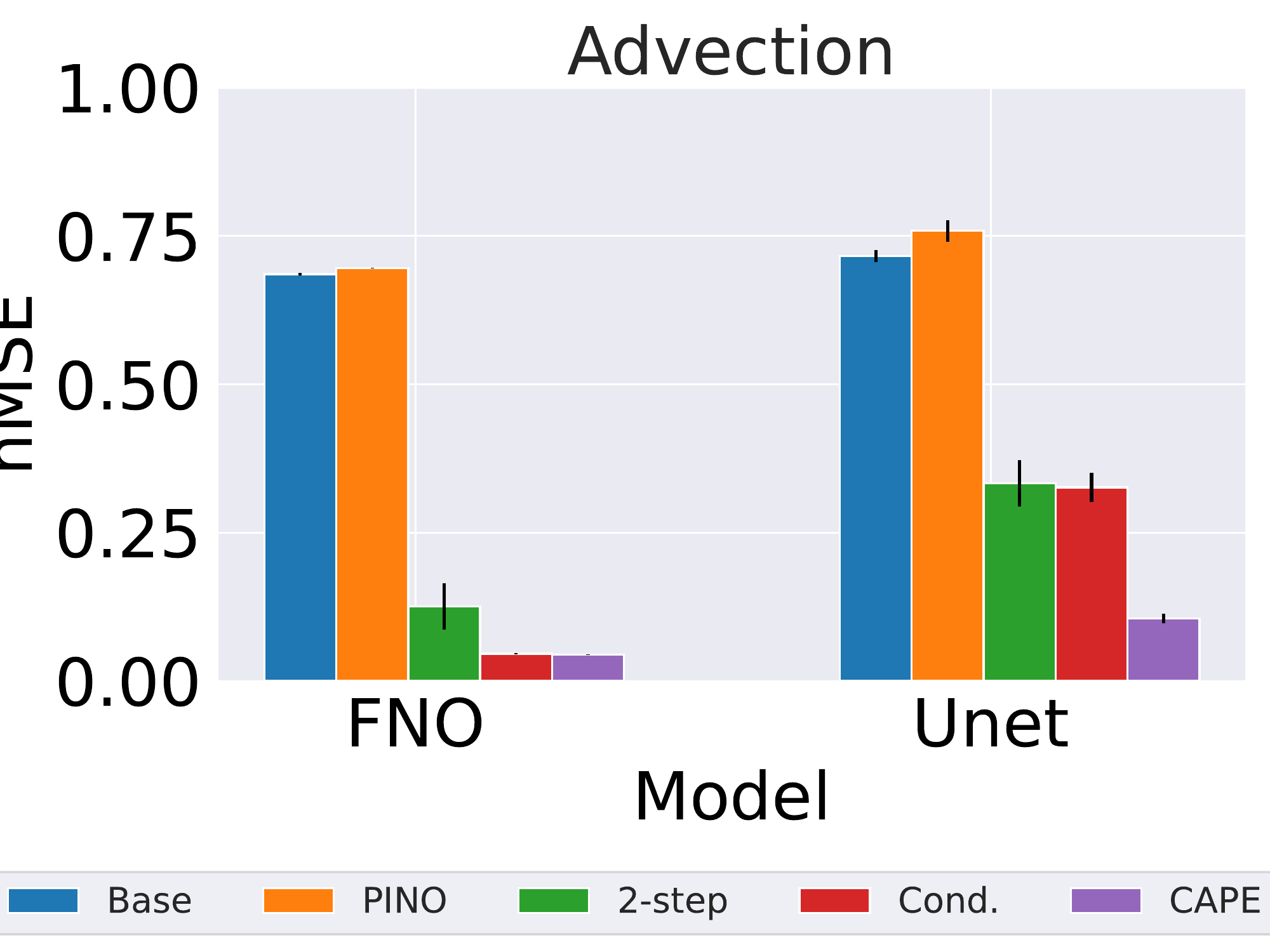}
    \includegraphics[width=0.31\textwidth]{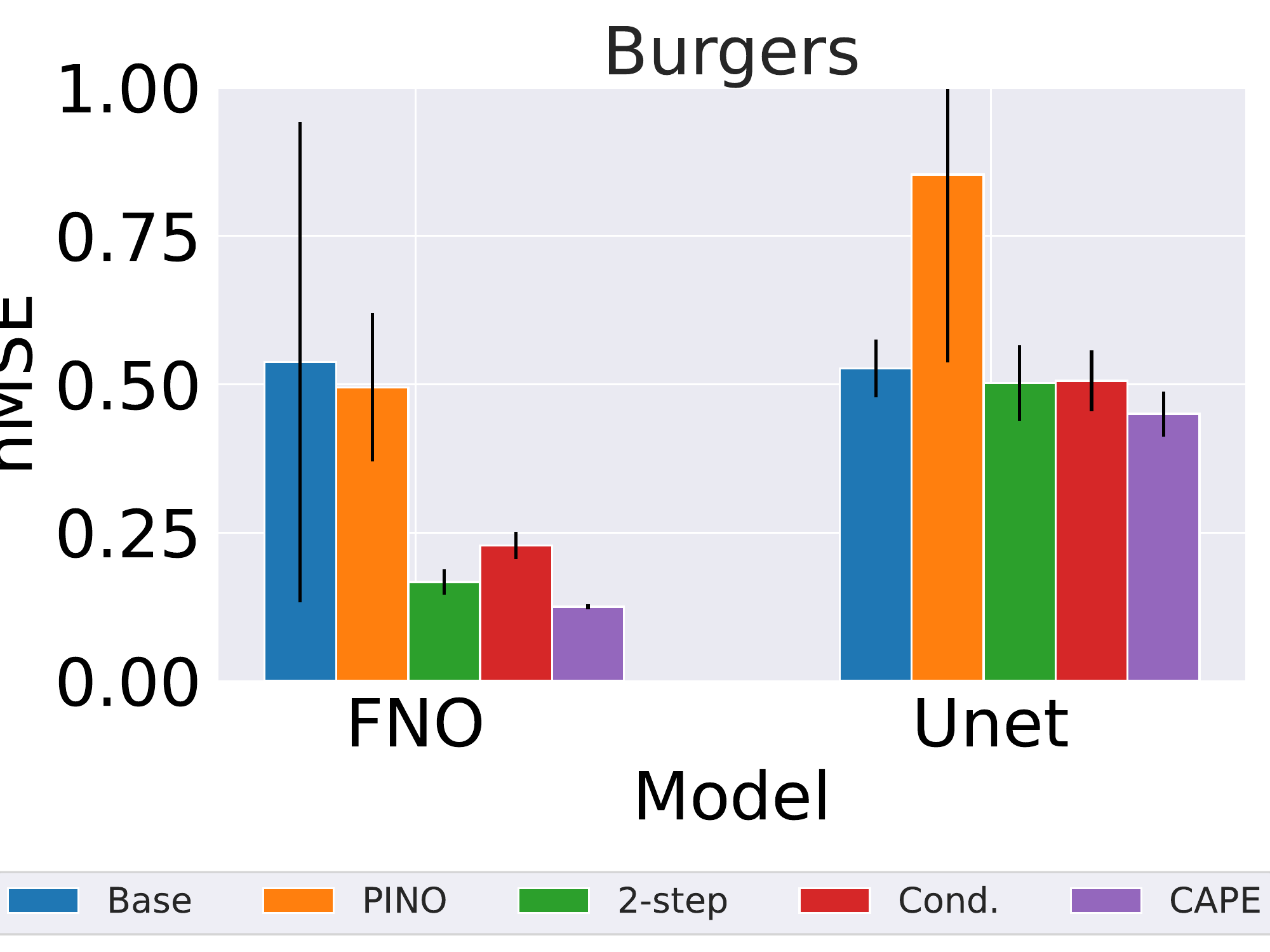}
    \includegraphics[width=0.31\textwidth]{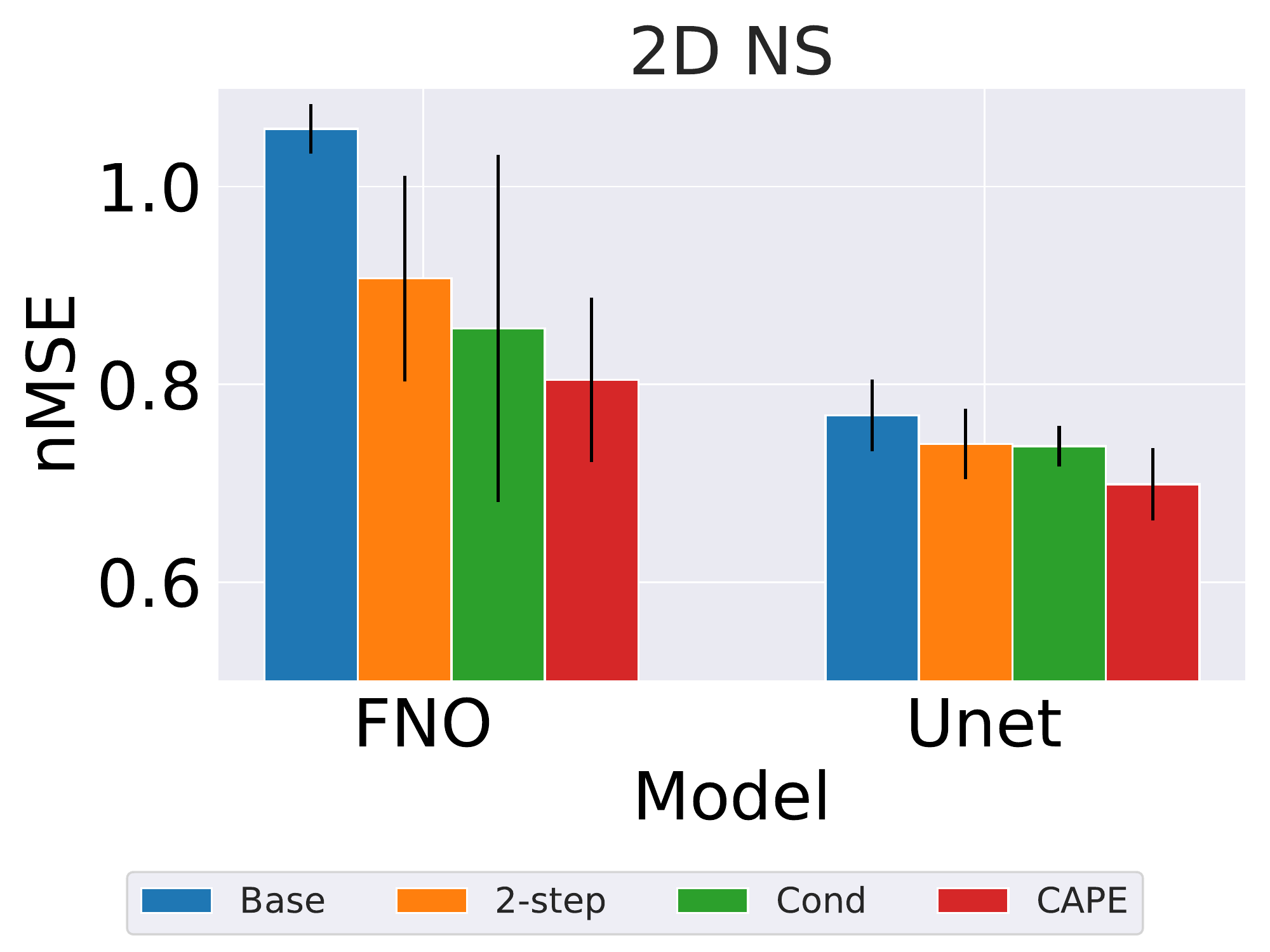}
     \caption{Plots of the normalized MSE (smaller is better) with an error bar for Advection eq. (Left), Burgers eq. (Middle), and 2D Compressible NS equations (Right).}
    \label{fig:Forward}
\end{figure*}

\begin{table*}[t]
\small
  \centering
  \begin{tabular}{lllllll}
  \toprule
    PDE & model &  \base &  \base (PINO) & Conditional & prev. 2-steps &  \cape  \\
  \midrule
  1D Advection & FNO &  $0.69^{\pm 2.2 \times 10^{-3}} $ & $0.70^{ \pm 1.6 \times 10^{-4}}$ & $0.05^{\pm 1.2 \times 10^{-3}}$ & $0.13^{\pm 3.9 \times 10^{-2}}$ & {\bf 0.03}$^{\pm 2.5 \times 10^{-3}}$ \\
          & Unet & $0.72^{\pm 1.0 \times 10^{-2}}$ & $0.76^{\pm 1.8 \times 10^{-2}}$ & $0.33^{\pm 2.0 \times 10^{-2}}$ & $0.33^{\pm 3.9 \times 10^{-2}}$ & {\bf 0.11}$^{\pm 8.3 \times 10^{-3}}$ \\
          & MPNN & $0.32^{\pm 2.5 \times 10^{-2}}$ & -- & $0.07^{\pm 2.0 \times 10^{-3}}$ & -- & -- \\          
  \midrule        
  1D Burgers & FNO & 0.54$^{\pm 0.40}$ & 0.49$^{\pm 1.3 \times 10^{-1}}$ & $0.23^{\pm 2.3 \times 10^{-2}}$ & 0.17$^{\pm 2.1 \times 10^{-2}}$ & {\bf 0.09}$^{\pm 7.8 \times 10^{-3}}$ \\
          & Unet & 0.53$^{\pm 4.9 \times 10^{-2}}$ & 0.85$^{\pm 3.2 \times 10^{-1}}$ & $0.51^{\pm 5.1 \times 10^{-2}}$ & 0.50$^{\pm 6.3 \times 10^{-2}}$ & {\bf 0.45}$^{\pm 3.8 \times 10^{-2}}$ \\
          & MPNN & $0.27^{\pm 0.11}$ & -- & $0.13^{\pm 3.7 \times 10^{-3}}$ & -- & -- \\
  \midrule        
  2D NS & FNO & 1.06$^{\pm 2.5 \times 10^{-2}}$ & -- & 0.86$^{\pm 0.18}$ & 0.91$^{\pm 0.10}$ & {\bf 0.80}$^{\pm 8.3 \times 10^{-2}}$ \\
        & Unet & 0.77$^{\pm 3.6 \times 10^{-2}}$ & -- & $0.74^{\pm 0.02}$ & 0.74$^{\pm 3.5 \times 10^{-2}}$ & {\bf 0.70}$^{\pm 3.7 \times 10^{-2}}$ \\
        & MPNN & N/A & -- & N/A & -- & -- \\        
        & TF-Net & N/A & N/A & N/A & $1.28^{\pm 0.19}$ & N/A \\                
  \bottomrule
  \end{tabular}
  \caption{List of the normalized RMSE (the smaller, the better) for Advection eq., Burgers eq., and 2D Compressible NS equations. }
  \label{tb:base_results}
\end{table*}  

%\textcolor{red}{%
%Noth that after applying numerical discretization of the above PDEs, the continuous variables $u(x,t), \rho(x,t), p(x,t), v(x,t)$ are expressed by the temporal-sequential tensor data: $\{ {\bf X}^{(t)}_{i,j,\cdots} \}$ as defined in \autoref{sec:prob_def} where the temporal coordinate $t$ is described by the temporal index $(t)$ and the spatial coordinate $x$ is described by the spatial index $i,j,\cdots$, which can be treated by the ML models. 
%}%
For 1-dimensional PDEs, we used $N=9000$ training instances and 1000 test instances for each PDE parameter with resolution 128. 
For 2-dimensional NS equations, we used $N=900$ training instances and 100 test instances for each PDE parameter with spatial resolution $64 \times 64$.

\textbf{Experiment Setup.}
We evaluated the neural models U-Net \citep{RonnebergerUNet2015} and FNO \citep{li2020fourier} with datasets provided by PDEBench \citep{PDEBench2022} for various  parameters for the 1D Advection equation, 1D Burgers equation, and 2D compressible Navier-Stokes equations. 
We also evaluated the message passing neural PDE Solvers (MPNN) \citep{brandstetter2022message} as a baseline allowing conditional treatment of PDE parameters.
%\textcolor{red}{%
  As a baseline for multi-dimensional turbulent flow, we evaluated the performance of the TF-Net \citep{wang2020towards} on 2D compressible Navier-Stokes equations. 
%}%
We trained each of the neural models (1) \textbf{Base}: without any changes (vanilla model), (2) \textbf{PINO}: with a PINO loss \citep{li2021physics}, (3) \textbf{Conditional}: the parameters are added to the input data as new channel-dimensions, (4) \textbf{2-step}: with the field data for the current and previous time-steps as input $(\bu^k, \bu^{k-1})$, and (5) \textbf{\textsc{CAPE}}: with the \cape module. 
%(3) AR method with $\ell=2$, and (4) with the \cape module. 
Other than case (4), we only provided field data for one  time step to the models and, therefore, the models cannot obtain PDE parameters' information from the given data. 
The PINO loss function regularizes the ML models to satisfy the residuals of the PDEs and might lead to an improved generalization behavior for unseen PDE parameters. 
The \cape module predicts intermediate field data for one future time step and this is used as the input to the \base model together with the field data of the current time step.\footnote{First introduced in \citep{li2020fourier} with twenty steps as input.}. 
The amount of field data provided to the \base network in cases (4) and (5) is the same: in case (4) the model always obtains field data for time steps $k$ and $k-1$ as input to predict the field data for time step $k+1$ while in case (5) the  \base model obtains field data for time step $k$ and intermediate field data for time step $k+1$ generated by the \cape model to predict the field data for time step $k+1$. Hence, the model in case (4) obtains two true field data for 2 initial steps and, therefore, has the opportunity to adapt to different PDE parameter values. Hence, while case (4) is more expensive, we consider it a strong baseline for the problem. 

Since the solutions of each PDE are not normalized 
%\footnote{Note that in general, a naive normalization can change the considering PDE form if the PDE is a non-linear one, e.g., $\partial_t (u/K) + (u/K)\partial_x (u/K) = (\partial_t u + (1/K) u\partial_x u)/K$.}
and based on prior results on evaluating PDE solvers \cite{PDEBenchDataset}, we measure the normalized RMSE (nRMSE).  %defined as 
% \begin{equation}
%   \mathrm{nMSE} \equiv \frac{||u_{\rm pred} - u_{\rm true}||_2}{||u_{\rm true}||_2},
% \end{equation}
%\begin{equation}
%  \mathrm{nMSE} \equiv \nicefrac{||u_{\rm pred} - u_{\rm true}||_2}{||u_{\rm true}||_2},
%\end{equation}
%where $||u||_2$ is the $L_2$-norm of a (vector-valued) variable $u$, and $u_{\rm true}, u_{\rm pred}$ are true and predicted value, respectively. 
%because it can be strongly affected by the mean value of the solution. 
We used the normalized RMSE loss function $L_{\rm nRMSE}$ with the auxiliary loss function of the \cape module $\mathbf{L}_{\rm cape} := \mathbf{L}_{\rm nMSE} + \alpha \mathbf{L}_{\rm cape}$ where $\alpha$ is the weight coefficient. 
The optimization was performed with Adam \citep{2015-kingma} for 100 epochs. The learning rate was divided by 2.0 every 20 epochs. 
For a fair comparison, we made the model size of the different methods as similar as possible. A table with model parameter sizes is provided in \autoref{tab:model_size} in the appendix. 
A more detailed description of the hyper-parameters is provided in \autoref{sec:setups} \footnote{
CAPE's hyper-parameters was determined using a smaller train-validation split, details of which are provided in \autoref{sec:hp_search}.}. 
%In the following experiments, we used various parameters of the PDEs for both of training and test sets. 
%Note that neither the vanilla FNO and U-net in principle cannot treat this task properly because they do not accept the PDE parameters. 

\begin{figure*}[!]
    \centering
    \includegraphics[width=0.19\textwidth]{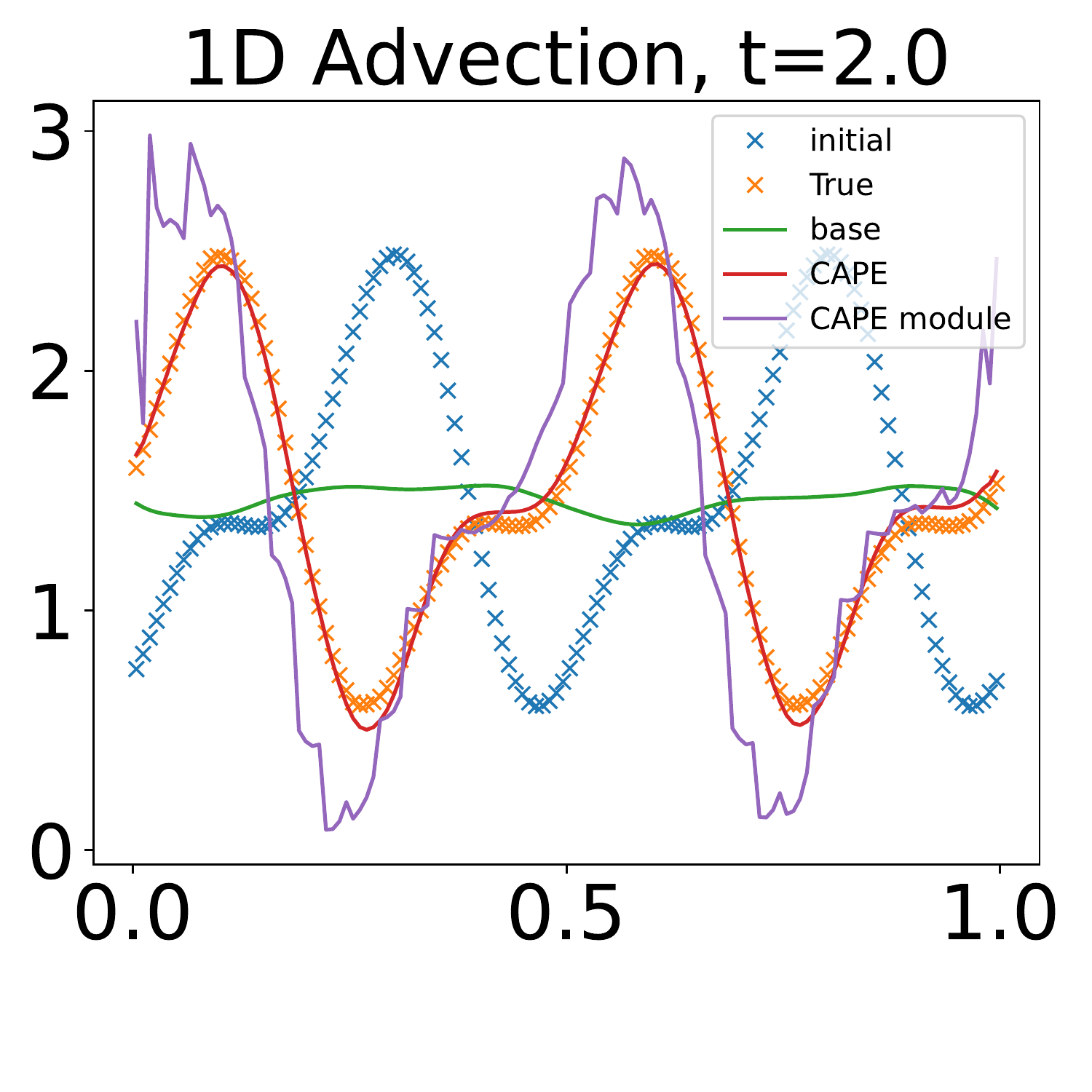}
    \includegraphics[width=0.19\textwidth]{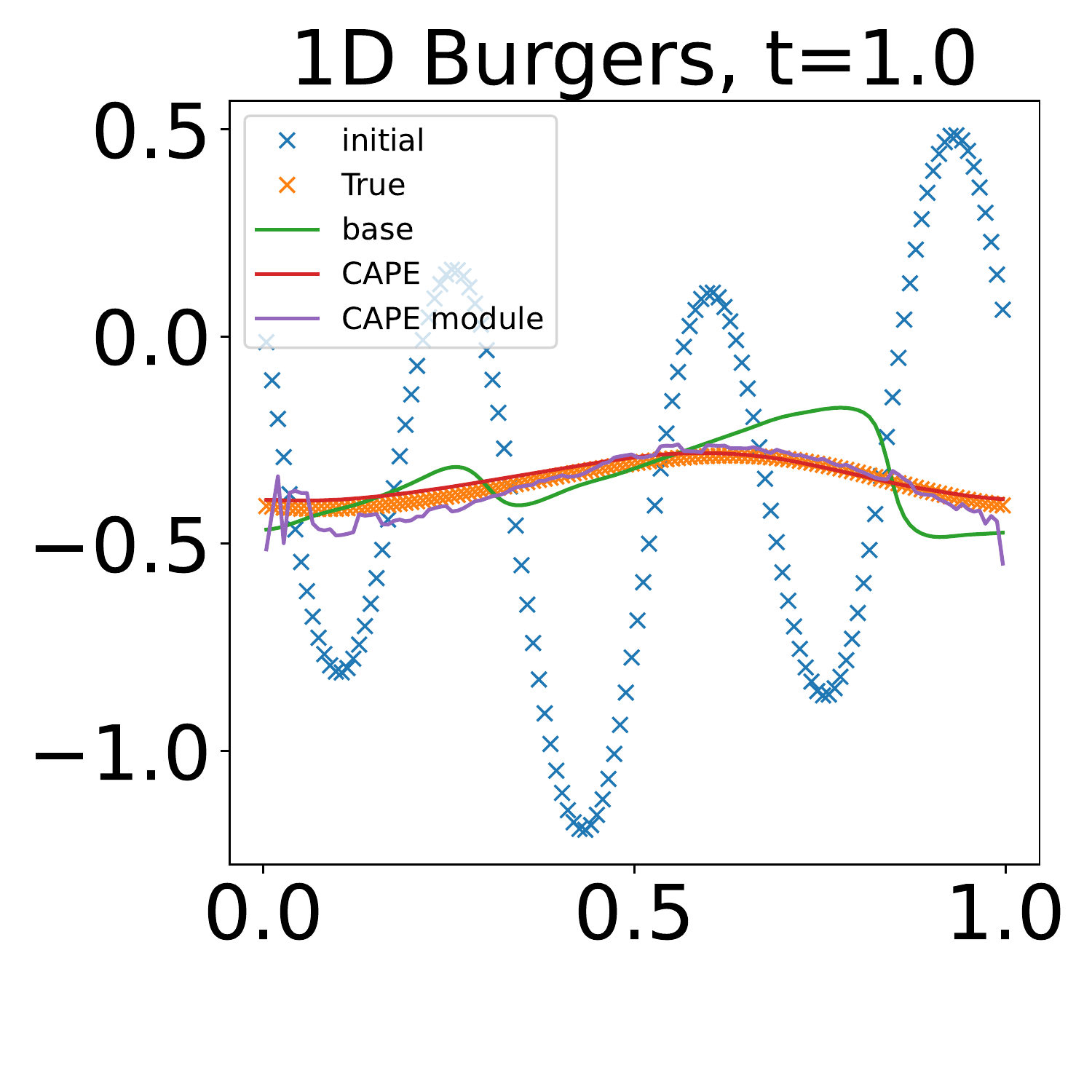}
    \includegraphics[width=0.6\textwidth]{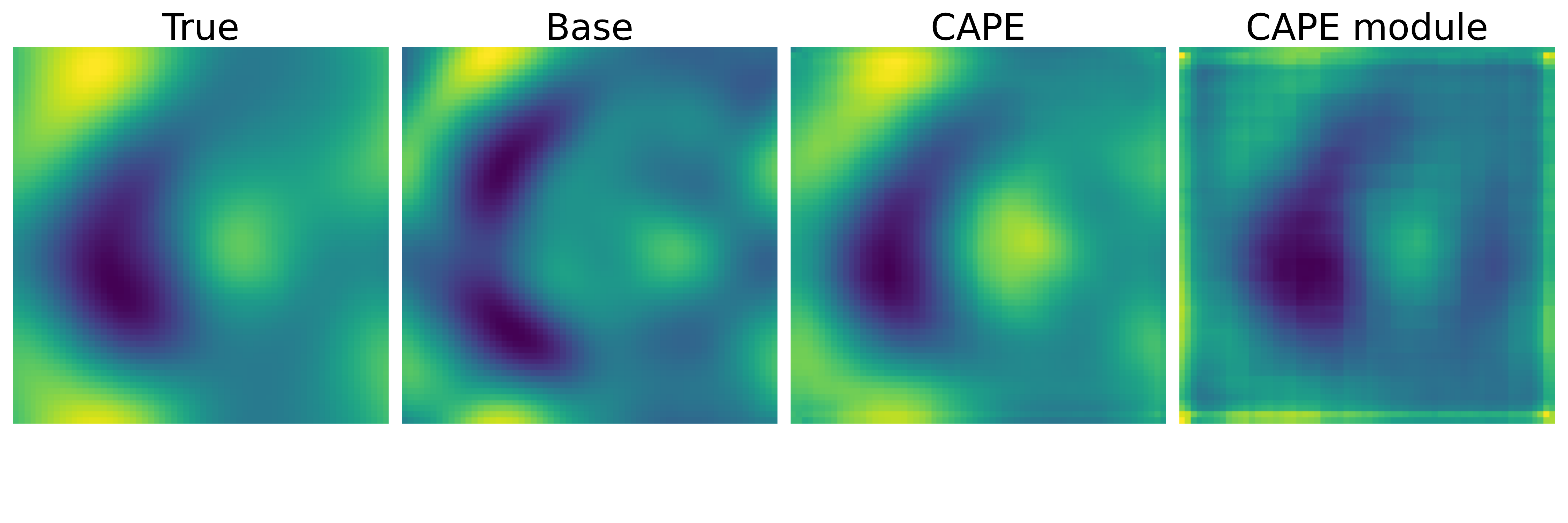}
     \caption{Visualization of the results: Advection eq. at the final time-step ($t=2.0$) (Left), Burgers eq. at $t_k = 20 (t=1.0)$ (2nd-left) at the final time-step, and $V_x$ of 2D NS equations at $t_k=5 (t=0.25)$ (Right). Here ``Base" is the vanilla FNO, ``\cape" is the FNO with \cape, ``\cape module" is the direct output from \cape module only; the \cape module provides a higher frequency proposal to the  \base model which then more accurately predicts the field data.}
    \label{fig:vis}
\end{figure*}

\textbf{Varying PDE parameters.}
\autoref{fig:Forward} shows bar plots comparing the  \base models with and without \cape module, the models with PINO loss, and the models with the 2-steps as input 
%\textcolor{red}{%
(see also \autoref{tb:base_results})\footnote{
%\textcolor{red}{%
The N/A for the results of 2D MPNN are because only 1D examples are provided in the official repository. Also TF-Net assumes to include more than one temporal steps, so the results other than "prev. 2-steps" becomes N/A.}
%}%
%}%
. The \cape module results in the lowest error in all cases. In particular, the \cape module leads to an impressible error reduction ranging from 20 \% (2D NS equation) to 95 \% (1D Advection). We partly attribute this to the  \base network's ability to capture physical dynamics from the PDE parameter-dependent data provided by the \cape module. The vanilla FNO is a state-of-the-art model and is superior to the U-net as a  \base model
\footnote{
  For the 2D NS PDE, the U-net achieves a smaller error than the FNO. We hypothesize that this is partly due to the difference in model size (the number of weights of the U-net is nearly 10 times larger) and partly because the U-net typically excels at image-to-image mapping problems. 
}. 
Interestingly, the PINO loss provides almost no benefit in our setting. We hypothesize that the PINO loss is heavily affected by and dependent on the time-step size. A more detailed explanation of this observation is given in \autoref{sec:PINO_discuss}. 
Interestingly, the \cape module provides either comparable or a little better results than the case with 2-step information. This indicates that the \cape module succeeded in providing equivalent and even more useful information to the  \base network.

\begin{figure*}[t!]
    \centering
    \includegraphics[width=0.24\textwidth]{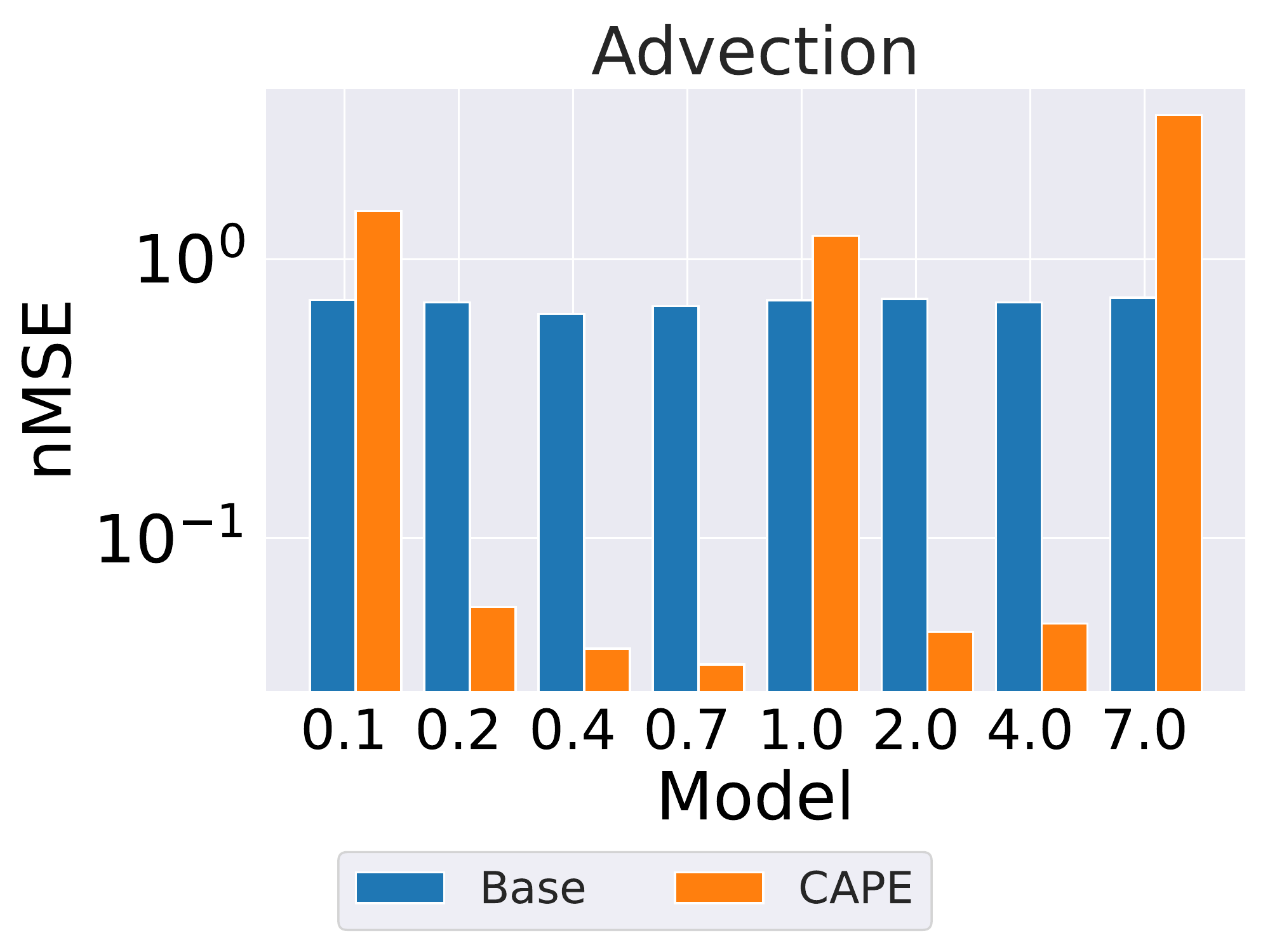}
    \includegraphics[width=0.24\textwidth]{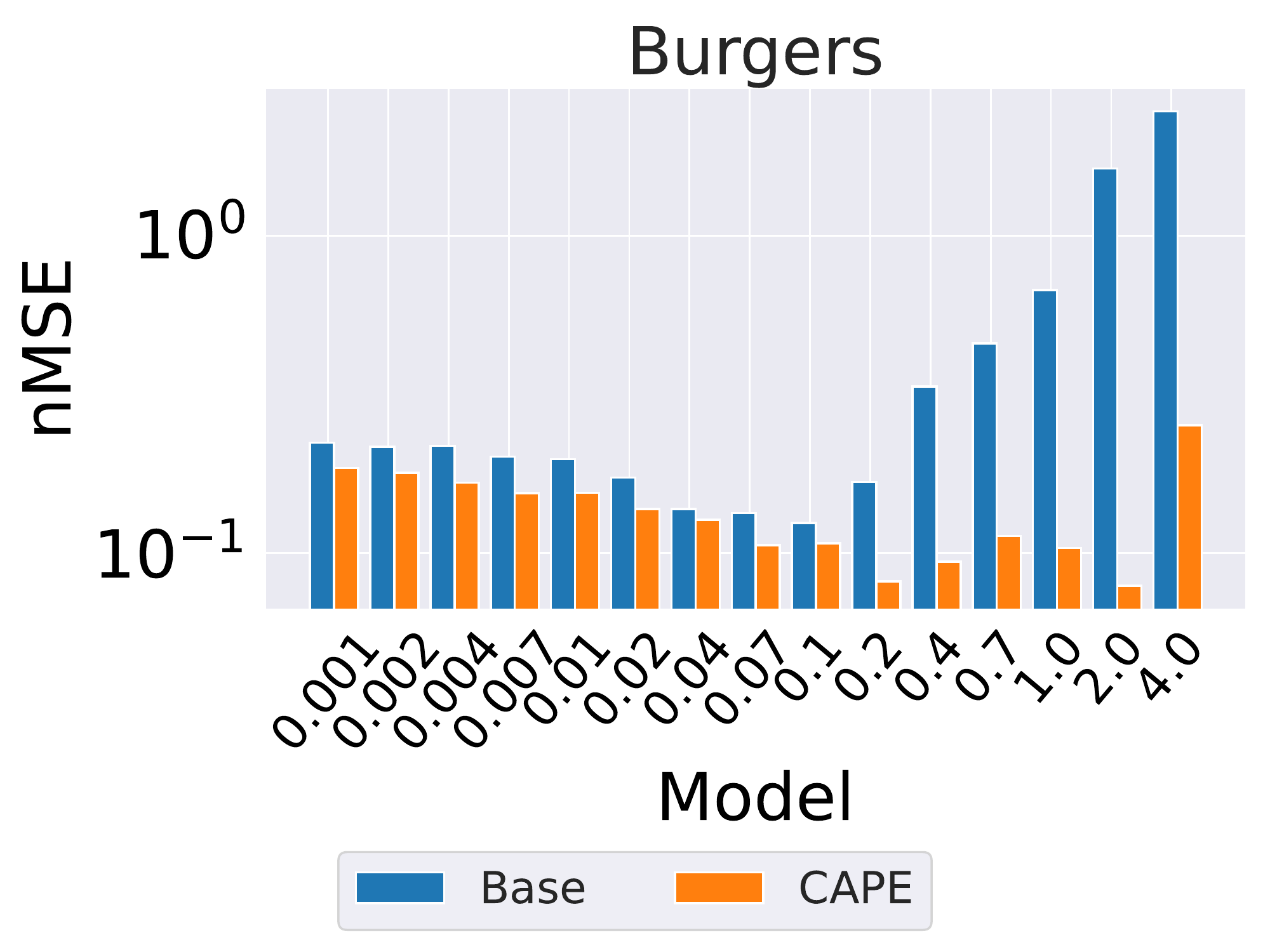}
    \includegraphics[width=0.24\textwidth]{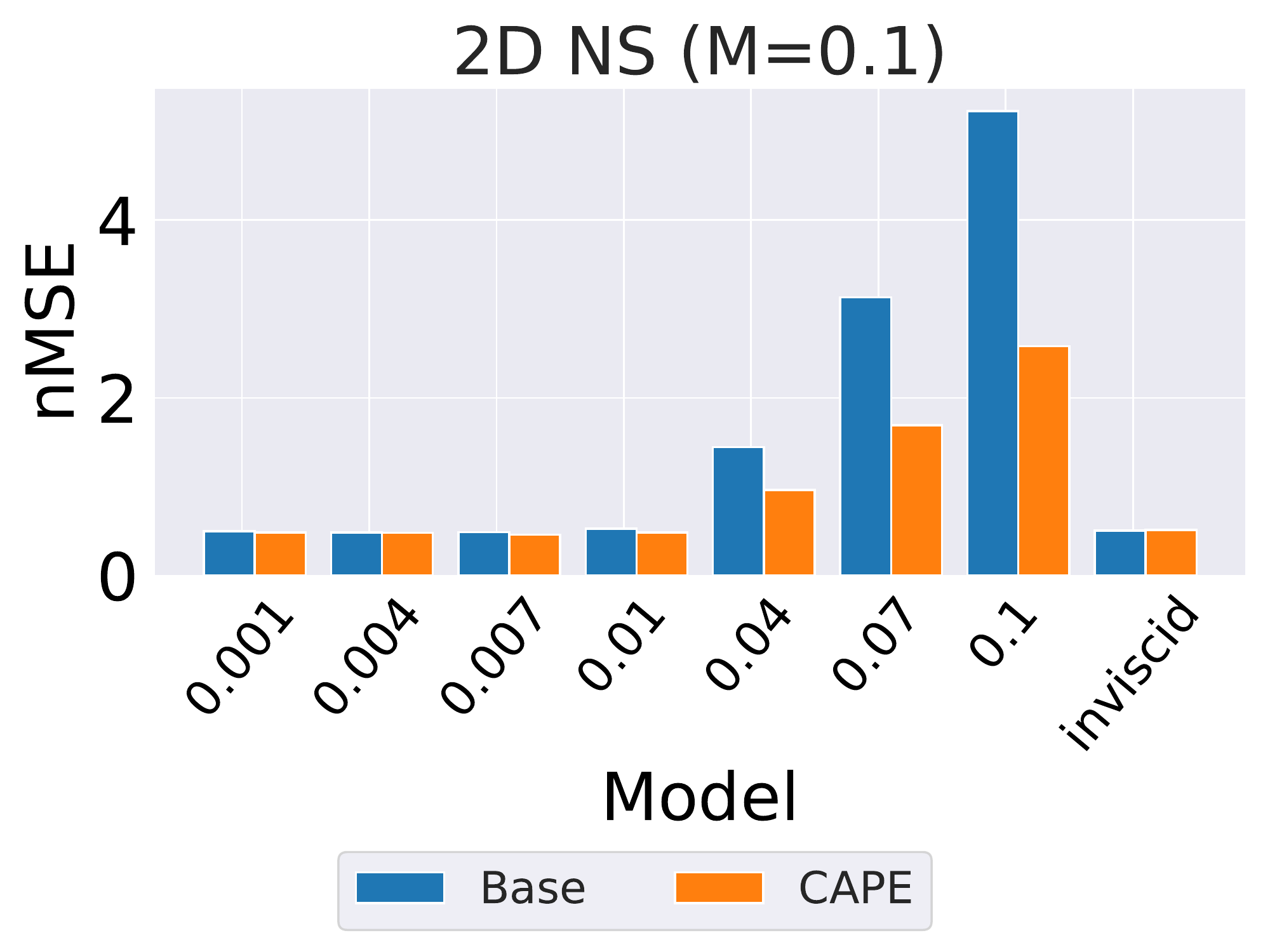}
    \includegraphics[width=0.24\textwidth]{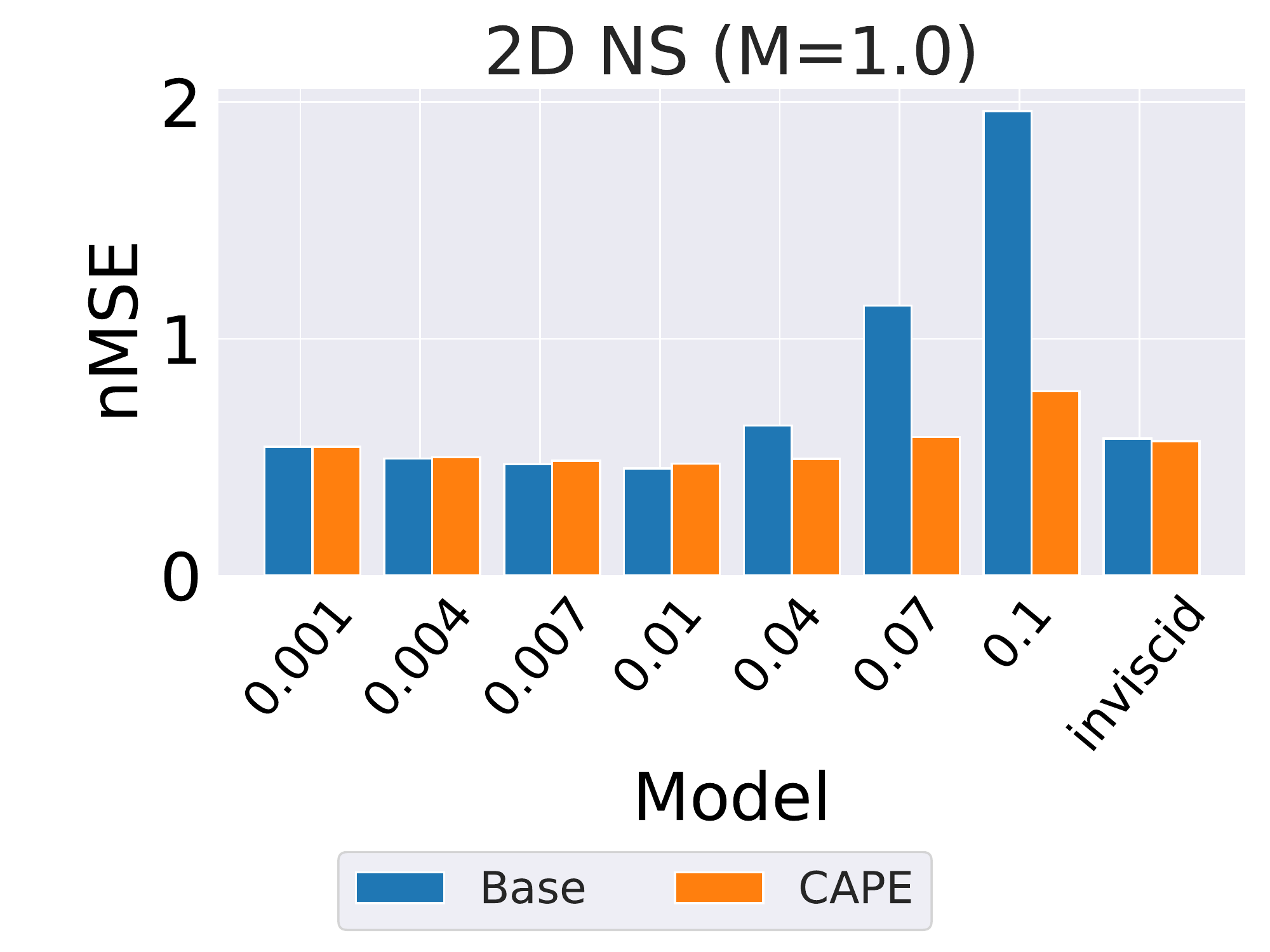}    
     \caption{Plots of the normalized MSE (smaller is better) in terms of each PDE parameter for Advection eq. (Left), Burgers eq. (Middle-Left), 2D NS with $M=0.1$ (Middle-right), and 2D NS with $M=1$ (Right). The results were obtained using the FNO.}
    \label{fig:Generalization}
\end{figure*}

\textbf{Generalization Ability.} \autoref{fig:Generalization} plots the normalized MSE for each parameter value of the PDEs using FNO as the  \base network. 
The parameter of the 1D Advection PDE controls the advection velocity and the parameters of the remaining equations control the strength of the diffusion process. 
First, the 1D-Advection result shows that the \cape module overfits with the trained parameter ($\beta = 0.2, 0.4, 0.7, 2.0, 4.0$), though it showed a very nice generalization performance to the trained PDE parameters. 
This could also be indicated by the fact that only the \cape result for the 1D Advection equation showed a much lower error than the approach receiving the 2-step field data which in theory provides a similar amount of information as the \cape module. %Hence, for a generalization to the unseen PDE parameters, we can use the difference of those errors as the smoking gun of the over-fitting of the \cape module to the trained parameters. %\mathias{please clarify the last sentence. Please do not use the word smoking gun :-)}
Hence, to evaluate generalization to the unseen PDE parameters, we expect the difference of these errors to be correlated with the \cape module overfitting to the trained parameters.

On the other hand, the other cases (the parameter describes the diffusion process) showed a good generalization to unseen PDE parameters. Note that the plots also indicate that vanilla models show a preference for the parameter regime; in all the cases, the vanilla models exhibit better results on smaller diffusion coefficients but lose accuracy as the diffusion coefficients increase. 

\begin{table}[t]
\small
  \centering
  \begin{tabular}{lll}
  \toprule
  Model & Ablation &  nMSE \\
  \midrule
  \multirow{4}{*}{FNO} & curriculum strategy & ${\bf 8.0\times10^{-1}}$ \\
  & fully autoregressive & $1.3\times10^{+0}$  \quad ($+ 0.5$) \\
  & only teacher-forcing & $3.2\times10^{+0}$  \quad (${\bf + 2.4}$) \\                  
  \midrule                  
  \multirow{4}{*}{Unet} & curriculum strategy & ${\bf 7.0\times10^{-1}}$ \\
  & fully autoregressive & $1.0\times10^{+0}$  \quad (${\bf + 0.3}$) \\
  & only teacher-forcing & $1.0\times10^{+0}$  \quad (${\bf + 0.3}$) \\                  
  \bottomrule
  \end{tabular}
  \caption{Ablation study for the 2D CFD equations with FNO and Unet as \base model.} %CS, AR, and TF means "curriculum stragety", "pure Autoregressive", and "pure Teacher-Forcing"}
   %Here AR is autoregressive and TR is teacher-forcing.}
  \label{tab:Ablation}
\end{table}

\textbf{Ablation experiments.} In this section, we performed an ablation study to separate the impact of the curriculum learning strategy. The ablation study on \cape structure is provided in \autoref{tab:ablation_CAPE_structure} in Appendix. 
\autoref{tab:Ablation} lists the result for the 2D NS equations using FNO and Unet as the \base network. 
The proposed curriculum learning strategy drastically impacts the accuracy of the model in all cases, indicating the effectiveness of seamlessly bridging teacher-forcing and auto-regressive training. %\textcolor{red}{In particular, we emphasize that our curriculum strategy is very important in the 2D NS case that is multi-dimensional and highly non-linear PDE, and that has actual  real-world applications.} 
The full ablation study results are provided in \autoref{tab:Ablation_full} in Appendix. 

\textbf{Qualitative analysis of the \cape module.}
In \autoref{fig:vis} we plot some representative outputs of the vanilla FNO, the \cape module, and the overall \cape model, and compare them with the true solutions. Interestingly, we can see that the  \base network often interpolates a higher noise approximation of the \cape module into the typical shape (style) of the final solution.

\textbf{Inference time.} We provide the comparison of the inference time between the hybrid approach (including 2 initial steps as input as done in the FNO paper) and \cape (only using the initial step as the input). Here we only consider a scenario where the initial time steps are obtained from a numerical simulation. \autoref{tab:inf_time} lists the inference time for solving the 2D NS equation. The inference time of the FNO with the \cape module is much shorter than the hybrid method where the inference time is dominated by the simulation time
\footnote{The experiments were run using an Nvidia GeForce RTX 3090 with CUDA-11.6. The ML models are implemented using PyTorch 1.12.1 and the numerical simulations with JAX-0.3.17.}.
%\footnote{This problem disappears when rich historic observational data could be accessible, which can provides the necessary several initial step data without numerical simulation.}.

\section{Related Work}
\label{others}

\paragraph{Scientific Machine Learning Models}
Scientific Machine Learning aims at data-driven modeling of physical systems. A notable example is Physics Informed Neural Networks \citep{2019JCoPh.378..686R,cai2022physics} (PINNs) that, having access to the PDE of a system, learns a neural network over the domain $\mathcal{X} \to \mathbb{R}^d$ by enforcing small residual, i.e. the error when the solution is evaluated by the PDE or the boundary conditions, over a set of sampled points. While PINNs have the capacity to model various physical systems, they need to be trained for each new condition or parameter. %\textcolor{red}{
After several pioneering work
%} 
\cite{long2018pde,long2019pde,wang2020towards,belbute2020combining}, Neural Operators \citep{li2021physics}, as FNO \citep{li2020fourier} or Graph NO \citep{li2020neural}, was proposed to model the continuous operators over an infinite space and have shown the ability to generalization at multiple scales. Also, more traditional image-to-image neural networks such as the U-Net \citep{RonnebergerUNet2015} can be adopted to model NOs.
Physics-Informed Neural Operators (PINO) \citep{li2021physics} improve the representational power of PINNs by pre-training a NO but having similar limitations to NOs. Message passing neural PDE solver \citep{brandstetter2022message} extends the message passing principle to solve PDEs. In signal processing, for the artificial bandwidth extension (ABE) task, the Time-Frequency Network\citep{dong2020time} (TFNet)  has been proposed, which shares a similar concept of channel attention.
% Unet, 
% FNO, 
% PINN
% TFNet, ...
% GNN base
\vspace{-.3cm}
\paragraph{Training Autoregressive Models} As was discussed in \autoref{sec:crc}, there are two representative training strategies for SciML, that is, teacher-forcing and auto-regressive training. The teacher-forcing strategy was originally developed in natural language processing  \citep{williams1989learning,NIPS2015_e995f98d} which predicts n+1-th step data (word) using the true n-th step information (word). This method is known to prevent from the error-accumulation in the predicted sequential data during the model training. In the case of Scientific ML, it was found that it is profitable to add a random noise for improving the robustness against the accumulated error at the inference time \citep{sanchez2018graph,sanchez2020learning,pfaff2020learning,stachenfeld2021learned}. On the other hand, the auto-regressive strategy uses the previous prediction of the model as n-th timestep information. Because of the error accumulation problem, not so many works were adopted, e.g. \citep{li2020fourier}. Recently, \citep{brandstetter2022message} proposed a reconciling method for this problem which is the so-called "pushforward trick". To increase stability, this method uses an adversarial-style loss which predicts the next timestep data using the previous prediction which is calculated using true data. Note that our method adopted a curriculum training strategy to prevent error accumulation instead of using an additional loss function. 
%auto-regressive
%teacher-forcing
% push-forward trick

\begin{comment}
   
\paragraph{Parameter Embedding} \francesco{I would remove this or move in the annex, it is not really related}
There has been an interest to put an additional information to DNN. For example,  Transformer-type models take into account the information of position of words in the sentence using positional encoding \citep{NIPS2017_3f5ee243,shaw2018self,huang2018music}. In the case of data generation, cGAN \citep{mirza2014conditional} accepts a conditional parameter to the generator network. In the case of SciML, PINN \citep{cai2022physics} and PINO \citep{li2021physics} can explicitly take into account PDE parameters during training but cannot change it during the test time. Recently  Message-passing PDE solver \citep{brandstetter2022message} was proposed in which PDE parameters and boundary conditions can be freely embedded into the network. However, this is specialized only for these models, and cannot apply the other models as our proposed method.      
\end{comment}

\begin{table}[!]
\small
    \centering
    \begin{tabular}{rrrr}
        \toprule
        PDE & Resolution  & Total Inference & simulation \\
        & & time [sec] & time [sec] \\        
        \midrule
        Simulation + FNO & $512^2$ & $582.8$ & $582.6$ \\
        \cape + FNO  & $512^2$ & $1.3$ & -- \\
        \bottomrule
    \end{tabular}
    %\end{tabularx}
    \caption{\label{tab:inf_time}Inference time comparison of simulation (initial steps=10) + FNO and \cape + FNO in the case of 2D CFD ($\eta=\zeta=0.1$). 
    The time-step size is $\Delta t = 0.05$ and the computations were performed until $t = 1.0$ as in this paper's other experiments.}   
\end{table}

\section{Conclusion and Limitations}

The \cape module allows any data-driven SciML models to incorporate PDE parameters. We  propose a simple but effective curriculum training strategy that allows us to bridge teacher-forcing and auto-regressive learning. We performed an extensive set of experiments and showed the effectiveness and efficiency of our method from various aspects: generalization of seen/unseen PDE parameters during training, parameter efficiency, and inference time. 

One of our key findings is the behavior of ML models without parameter embeddings which either (1) exhibit poor performance uniformly for all the PDE parameters (1D Advection eq.), or (2) overfit to a specific parameter regime (1D Burgers eq. and 2D NS eqs.). Hence, the ML model in this case does not generalization to PDE parameters unseen during training. On the other hand, the models with \cape module generalized well for 1D Burgers' eq. and 2D NS eqs. whose parameters govern the physical systems' diffusion behavior. \cape cannot be generalized for 1D Advection eq., and we consider it necessary to formulate \cape to be more physics-informed which we consider for future work. 

\paragraph{Limitations}
% CAPE method is applicable only to field equation type problem, such as solving hydrodynamic equation, so it cannot be applied to particle simulations as the molecular dynamic simulation. Also CAPE in this paper is aiming to CNN base model. Applying to GNN method is our future work. 
\cape's scope is restricted to  classical field equation problems such as solving hydrodynamic equations. It cannot be applied to particle simulations such as molecular dynamics simulations. Since \cape is using convolution attention, it is limited to regular grids. 
% Applying to GNN method is our future work. 

%\begin{comment}

\subsubsection*{Acknowledgments}
%Use unnumbered third-level headings for the acknowledgments. All
%acknowledgments, including those to funding agencies, go at the end of the paper.
We thank Marimuthu Kalimuthu for many fruitful comments on our manuscript. We acknowledge support by the Stuttgart Center for Simulation Science (SimTech). 
%\end{comment}

\clearpage

\begin{comment}
% Acknowledgements should only appear in the accepted version.
\section*{Acknowledgements}

\textbf{Do not} include acknowledgements in the initial version of
the paper submitted for blind review.

If a paper is accepted, the final camera-ready version can (and
probably should) include acknowledgements. In this case, please
place such acknowledgements in an unnumbered section at the
end of the paper. Typically, this will include thanks to reviewers
who gave useful comments, to colleagues who contributed to the ideas,
and to funding agencies and corporate sponsors that provided financial
support.

% In the unusual situation where you want a paper to appear in the
% references without citing it in the main text, use \nocite
\nocite{langley00}
\end{comment}
\bibliography{ICML2023}
\bibliographystyle{icml2023}

%%%%%%%%%%%%%%%%%%%%%%%%%%%%%%%%%%%%%%%%%%%%%%%%%%%%%%%%%%%%%%%%%%%%%%%%%%%%%%%
%%%%%%%%%%%%%%%%%%%%%%%%%%%%%%%%%%%%%%%%%%%%%%%%%%%%%%%%%%%%%%%%%%%%%%%%%%%%%%%
% APPENDIX
%%%%%%%%%%%%%%%%%%%%%%%%%%%%%%%%%%%%%%%%%%%%%%%%%%%%%%%%%%%%%%%%%%%%%%%%%%%%%%%
%%%%%%%%%%%%%%%%%%%%%%%%%%%%%%%%%%%%%%%%%%%%%%%%%%%%%%%%%%%%%%%%%%%%%%%%%%%%%%%
\newpage
\appendix
\onecolumn

\section*{Learning Neural PDE Solvers with Parameter-Guided Channel Attention}
%\section*{CAPE: Channel-Attention-Based PDE Parameter Embeddings for SciML - Supplementary Material}

\section{Additional Related Work}
\paragraph{Parameter Embedding} %\francesco{I would remove this or move in the annex, it is not really related}
There has been an interest to put additional information to DNN. For example,  Transformer-type models take into account the information of the position of words in the sentence using positional encoding \citep{NIPS2017_3f5ee243,shaw2018self,huang2018music}. In the case of data generation, cGAN \citep{mirza2014conditional} accepts a conditional parameter to the generator network. In the case of SciML, PINN \citep{cai2022physics} and PINO \citep{li2021physics} can explicitly take into account PDE parameters during training but cannot change them during the test time. Recently  Message-passing PDE solver \citep{brandstetter2022message} was proposed in which PDE parameters and boundary conditions can be freely embedded into the network. However, this is specialized only for these models, and cannot apply the other models as our proposed method.  

\section{\label{sec:setups}Detailed Training Setup}

\subsection{General Setup}

As is explained in \autoref{sec:experiments}, 
we used datasets provided by PDEBench \citep{PDEBenchDataset} a benchmark for SciML from which we downloaded datasets of the following PDEs: 1D Advection equation, 1D Burgers equation, and 2D compressible NS equations. 
For 1-dimensional PDEs, we used $N=9000$ training instances and 1000 test instances for each PDE parameter with spatial resolution: 128 ($\Delta x = 1/128$) and temporal step-size: $\Delta t = 0.05$. 
For 2-dimensional NS equations, we used $N=900$ training instances and 100 test instances for each PDE parameter with spatial resolution: $64 \times 64 \ (\Delta x = \Delta y = 1/128)$ and temporal step-size: $\Delta t = 0.05$. Smaller temporal step-size results are also provided in \autoref{sec:larger-timestep}. 

Concerning the training, 
the optimization was performed with Adam \citep{2015-kingma} for 100 epochs. The learning rate was set as $3 \times 10^{-3}$ which is divided by 2.0 every 20 epochs. 
The mini-batch size we used was 50 for all the cases. 
To stabilize the \cape module's training in the initial phase, we empirically found it is a little better if we have a warm-up phase during which only \cape module is updated. We performed warm-up for the first 3 epochs, which slightly reduce the final performance fluctuations resulting from the randomness of the initial weights of the network. In the \cape module, the kernel size of the depth-wise convolution was set as: 5. %In our experiments, we stacked 3 \cape modules before providing the output with the \base networks. Note that the channel parameter $c$ of the second and 3rd network were set as: $2 c$. 
The training was performed on GeForce RTX 2080 GPU for 1D PDEs and GeForce GTX 3090 for 2D NS equations. 
For PINO loss, we set the coefficient $1$ following the original implementation. 

\subsection{Hyper-parameter Selection\label{sec:hp_search}}

The hyper-parameters and the \base network parameters are listed in \autoref{tab:network-parameters}. 
%\textcolor{red}{%
To clarify that our result with \cape was not overfitted to the test dataset, we also performed a hyper-parameter search of the coefficient $\alpha$ of the loss term for \cape $L_{\rm cape}$ which is the unique hyper-parameter we can tune in the experiments. 
We created new data for 1D Advection equation with advection velocity $\beta = 0.3, 0.5, 1.2$ which are not included in our main experiments provided in \autoref{tab:PDE_params}. We split the data into train/validation/test with ratio: $(0.9,0.05,0.05)$, and saved the best model in terms of the validation loss value. The results are summarized in \autoref{tab:Validation_results} which indicates that the best parameter exists around $\alpha = 10^{-4}$ independent of the test set, and our choice in \autoref{tab:network-parameters} is validated. 
%}%

% \subsubsection{ {\color{red} Networks' sizes comparison}}
% \label{sec:size}
% The networks' structures of the \base models is presented in \autoref{tab:network-parameters}, while the resulting network size is listed in \autoref{tab:model_size}. 

\begin{table}[h]
  \centering
  \begin{tabular}{llrrrrr}
  \toprule
  Dimension & Model & width & mode & d & mode (\cape)& $\alpha$ \\
  \midrule
 \multirow{3}{*}{1D} & FNO & 36 & 12 & -- & -- & -- \\
                  & FNO w.t. \cape & 20 & 12 & 64 & 12 & $5.7 \times 10^{-5}$ \\
  \midrule
 \multirow{3}{*}{2D} & FNO & 28 & 12 & -- & -- \\
                  & FNO w.t. \cape & 20 & 12 & 64 & 9 & $8.3 \times 10^{-5}$ \\
  \toprule
  Dimension & Model & init features &  & d & mode (\cape)& $\alpha$ \\
  \midrule
 \multirow{3}{*}{1D} & Unet & 32 &  & -- & -- & -- \\
                  & Unet w.t. \cape & 32 &  & 64 & 12 & $5.7 \times 10^{-5}$ \\
  \midrule
 \multirow{3}{*}{2D} & Unet & 32 &  & -- & -- \\
                  & Unet w.t. \cape & 30 &  & 64 & 9 & $8.3 \times 10^{-5}$ \\
  \bottomrule
  \end{tabular}
  \caption{Network Parameters. Where $d$ is the channel number of CNNs in \cape.}
  \label{tab:network-parameters}
\end{table}

\begin{table}[h]
  \centering
  \begin{tabular}{lrr}
  \toprule
  PDE & $\alpha$ &  nRMSE \\
  \midrule
 \multirow{3}{*}{1D Advection} & $10^{-6}$ & 0.069 \\
                  & $10^{-5}$ & 0.073  \\
                  & $5 \times 10^{-5}$ & 0.046  \\     
                  & $10^{-4}$ & 0.045  \\                  
                  & $10^{-3}$ & 0.068 \\
  \bottomrule
  \end{tabular}
  \caption{Results on Train/Validation/Test data set with 1D Advection equation. The model is FNO with \cape.}
  \label{tab:Validation_results}
\end{table}

\subsection{ {Networks' sizes comparison}}
\label{sec:size}
The networks' structures of the \base models are presented in \autoref{tab:network-parameters}, while the resulting network size is listed in \autoref{tab:model_size}. 

\begin{table}[h]
  \centering
  \begin{tabular}{llr}
  \toprule
  Dimension & Model &  \# Parameters \\
  \midrule
 \multirow{3}{*}{1D} & FNO & 73K \\
                  & FNO w.t. \cape & 68K  \\
                  & Unet & 2.71M \\
                  & Unet w.t. \cape & 2.75M \\ 
                  & MPNN & 614k \\                   
  \midrule
 \multirow{3}{*}{2D} & FNO & 0.91M \\
                  & FNO w.t. \cape & 0.82M \\
                  & Unet & 7.8M\\
                  & Unet w.t. \cape & 7.2M \\
                  & TF-net & 7.4M\\
  \bottomrule
  \end{tabular}
  \caption{Model Size}
  \label{tab:model_size}
\end{table}

\subsection{Conditional Modelling}

Here we provide a more detailed explanation for the conditional models in \autoref{sec:experiments}. 
In this paper, the conditional models have the same model structures as the vanilla ones, but we only change the input data as: 
\begin{equation}
  u^k \in \mathbb{R}^{C \times N_1 \times ...} \rightarrow 
  \mathrm{concatenate}(u^k, \lambda) \in \mathbb{R}^{(C+1) \times N_1 \times ...}, 
\end{equation}
where the PDE parameter are taken as a part of the input by concatenating it to the field data’s new channel dimension. 
Although it is possible to consider a more elaborate method, such as performing an MLP on the PDE parameters, we avoid those cases for simplicity. 

\subsection{Modification on the Message-Passing PDE Solvers}

In \autoref{sec:experiments}, we consider the Message-Passing PDE Solvers \citep{brandstetter2022message} as a baseline model that accepts PDE parameters. For a fair comparison, we are forced to modify the model as 
(1) accepting only 1-time step data, (2) adding a case of "time-window" parameter with 10 for the decoder. 
Concerning the first case, the original model assumes to accept sequential data whose time-step size must be equal to the size of the "time-window" parameter. 
For the second modification, we added a new 1D-convolution layer accepting the "time-window" parameter equal to 10. The detailed structure of the new decoder is provided in \autoref{tab:Dec_CNN_MPNN}.

\begin{table}[h]
  \centering
  \begin{tabular}{lrrrr}
  \toprule
  Module & in-channel & out-channel & kernel size & stride \\
  \midrule
  1D Conv-1 & 1 & 8 & 18 & 5 \\
  1D Conv-2 & 8 & 1 & 14 & 1 \\  
  \bottomrule
  \end{tabular}
  \caption{Decoder CNN structure for Message-Passing PDE Solvers}
  \label{tab:Dec_CNN_MPNN}
\end{table}

%\color{red}
\subsection{Modification on the TF-Net\label{sec:mod-tfnet}}

In \autoref{sec:experiments}, we employed the TF-Net \citep{wang2020towards} as a baseline model for 2D compressible Navier-Stokes equations. 
However, the original model does not allow us to flexibly change the temporal filter size (the number of temporal time-step). Consequently, we made modifications to the model to suit our experimental setup, where only previous 2 time-step data was considered. 
In our experiment, we replaced the original temporal filter into a simple $1 \times 1$-Convolution layer whose kernel size is unity and the channel number is two for both input and output channels. 
We consider that the somewhat unsatisfactory results obtained with our TF-Net can be partly attributed to this modification. %and does not reflect the original model performance perfectly. 

\section{An additional experiments on 2D Burgers equation\label{sec:2d-burgers}}

In this section, we provide an additional experiment results conducted on 2-dimensional Burgers equation whose expression is given as:  
\begin{align}
    \partial_t u(t,x,y) + u(t,x,y) (\partial_x u(t,x,y) + \partial_y u(t,x,y)) &= \nu/\pi (\partial_{xx} u(t,x,y) + \partial_{yy} u(t,x,y)),
    \label{eq:bgs-2d}
\end{align}
where $\nu$ is the diffusion coefficient and the parameter of this equation. 
Note that we keep the variable as a scalar function and also the diffusion coefficient as a constant for both of the spatial directions, for simplicity. The considered diffusion coefficient are listed in \autoref{tab:PDE_params-2dburgers}. The results are provided in \autoref{tb:base_results-2dburgers} which shows our \cape module provides better results than the conditional model\footnote{The experimental setup adheres to the configuration employed in the 2D CFD scenario.}. 

\begin{table*}[t]
\small
  \centering
  \begin{tabular}{llllll}
  \toprule
    PDE & model &  \base & Conditional & prev. 2-steps &  \cape  \\
  \midrule
  2D Burgers & FNO & 0.0357$^{\pm 0.8 \times 10^{-3}}$ & 0.0274$^{\pm 0.4 \times 10^{-3}}$ & {\bf 0.0249}$^{\pm 1.5 \times 10^{-3}}$ & 0.0267$^{\pm 2.6 \times 10^{-3}}$ \\
          & Unet & 0.31$^{\pm 0.04}$ & $0.32^{\pm 0.10}$ & {\bf 0.21}$^{\pm 0.07}$ & 0.28$^{\pm 0.06}$ \\
          & TFNet & N/A & N/A & $0.0427^{\pm 0.023}$ & N/A \\
  \bottomrule
  \end{tabular}
  \caption{List of the normalized RMSE (the smaller, the better) for 2D Burgers eq.}
  \label{tb:base_results-2dburgers}
\end{table*}  

\begin{table*}[t]
\small
\centering
    \begin{tabular}{ll}
        \toprule
        PDE & training \& test parameters \\
        \midrule
        2D Burgers & $\nu = (0.001, 0.004, 0.01, 0.04, 0.1, 0.4, 1.0)$ \\
        \bottomrule
    \end{tabular}
    \caption{PDE parameters used in the experiments on 2D Burgers equation.\label{tab:PDE_params-2dburgers}}
    \vspace{-0.2cm}
\end{table*}

%\color{black}

\section{Discussion of Results for the PINO Loss}
\label{sec:PINO_discuss}

\paragraph{Reason why PINO loss does not work}

The PINO loss function is an emulation of the PINO loss function using ML's output. 
For example, the PINO loss function of the 1D Advection equation case is:
\begin{equation}
    L_{\rm PINO} = \frac{u^{n+1}_j - u^{n-1}_j}{2 \Delta t} - \beta \mathcal{F}^{-1} (i k \mathcal{F}(u)).
    \label{eq:PINO}
\end{equation}

On the other hand, the usual spectral method solves the equation as: 
\begin{equation}
    \tilde{u}^{n+1}_j = u^{n-1}_j + 2 \Delta t \beta \mathcal{F}^{-1} (i k \mathcal{F}(u)).
    \label{eq:Spect}
\end{equation}
By substituting \autoref{eq:Spect}, \autoref{eq:PINO} reduces to: 
\begin{equation}
    L_{\rm PINO} = \frac{u^{n+1}_j - \tilde{u}^{n+1}_j}{2 \Delta t}.
    \label{eq:PINO_mod}
\end{equation}
This shows that the PINO loss function penalizes the machine learning model prediction to be close to the spectral method prediction. 
However, in general, the classical direct simulation methods have to use the time-step size $\Delta t$  restricted by the theoretical stability condition, such as the CFL condition. 
And the prediction $\tilde{u}^{n+1}_j$ becomes completely wrong if the used $\Delta t$ does not satisfy the stability condition, resulting in the PINO loss function leading to a completely harmful effect for the ML models. 
%which is not taken into account for estimating $\{ u^n \}$ via ML models. 
%This indicates that \autoref{eq:PINO} does not becomes small even if $u^{n+1}$ is close to the real solution because the equation becomes $0$ only when $u^{n+1} = \tilde{u}^{n+1}$ in \autoref{eq:Spect}. 
In our experiments, $\Delta t = 0.05$ is larger than the time step demanded by the stability condition, e.g., $\Delta t < \Delta x / \beta  \sim 0.0025$ if we set $\Delta x = 1/128, \beta = 1$. 
So, we consider that our experiment result showing worse error from PINO loss function is a natural result from this consideration.  

\section{Detailed Results}

\subsection{PDE Parameter Dependence Study}

In \autoref{tab:pde-param-adv}, \autoref{tab:pde-param-bgs}, and \autoref{tab:pde-param-2DNS}, we provide the nRMSE values plotted in \autoref{fig:Generalization}. 

\begin{table}[h]
  \centering
  \begin{tabular}{llrrrrrrrr}
  \toprule
     & $\beta$ & 0.1 & 0.2 & 0.4 & 0.7 & 1.0 & 2.0 & 4.0 & 7.0 \\
  PDE & type &  &    &    &    &  &    &   &  \\
  \midrule
  Advection &  \base & 0.716 & 0.700 & 0.638 & 0.680 & 0.714 & 0.721 &  0.700 & 0.729 \\
            & \cape & 0.846 & 0.056 & 0.040 & 0.035 & 1.218 & 0.046 & 0.049 & 3.300 \\
\bottomrule            
  \end{tabular}        
  \caption{The nRMSEs FNO with \cape in terms of each advection velocity of 1D Advection equation. Visualization is given in \autoref{fig:Generalization}}
  \label{tab:pde-param-adv}
\end{table}  

\begin{table}[h]
\centering
%\begin{tabularx}{\textwidth}{llrrrrrrrrr}
\begin{tabular}{llrrrrrrrrr}
%\begin{tabularx}{llrrrrrrrrr}
%\resizebox{\linewidth}{!}{\begin{tabular}{llrrrrrrrrr}
  \toprule
     & $\nu$ & 0.001 & 0.002 & 0.004 & 0.007 & 0.01 & 0.02 & 0.04 & 0.07 & 0.1 \\
  PDE & type &  &    &    &    &  &    &   & &  \\     
    \midrule  
  Burgers &  \base & 0.223 & 0.216 & 0.218 & 0.201 & 0.198 & 0.173 & 0.138 & 0.134 & 0.124 \\
          & \cape & 0.185 & 0.179 & 0.167 & 0.155 & 0.155 & 0.138 & 0.127 & 0.106 & 0.107  \\
\bottomrule          
  \end{tabular}
%}  
  \caption{The nRMSEs FNO with \cape in terms of each diffusion coefficient of 1D Burgers equation. Visualization is given in \autoref{fig:Generalization}}
  \label{tab:pde-param-bgs}
\end{table}

\begin{table}[h]
  \centering
  \begin{tabular}{llrrrrrr}
  \toprule
     & $\nu$ &  0.2 & 0.4 & 0.7 & 1.0 & 2.0 & 4.0 \\
  PDE & type & & & & & & \\     
    \midrule  
  Burgers &  \base & 0.168 & 0.335 & 0.458 & 0.674 & 1.626 & 2.460 \\
          & \cape & 0.081 & 0.094 & 0.113 & 0.104 & 0.079 & 0.253 \\
\bottomrule          
  \end{tabular}
\end{table}  

\begin{table}[h]
  \centering
  \begin{tabular}{llrrrrrrrr}
  \toprule
     & $\eta=\zeta$ & $10^{-8}$ & 0.001 & 0.004 & 0.007 & 0.01 & 0.04 & 0.07 & 0.1 \\
  PDE & type &  &    &    &    &  &    &   &  \\
    \midrule
    2D NS ($M=0.1$) &  \base & 0.508 & 0.500 & 0.488 & 0.491 & 0.529 & 1.447 & 3.132 & 5.228  \\
    & \cape & 0.516 & 0.487 & 0.482 & 0.462 & 0.486 & 0.965 & 1.692 & 2.582 \\
    \midrule          
    2D NS ($M=1.0$)&  \base & 0.579 & 0.545 & 0.495 & 0.471 & 0.453 & 0.635 & 1.141 & 1.962  \\
    & \cape & 0.569 & 0.544 & 0.501 & 0.485 & 0.474 & 0.494 & 0.585 & 0.779 \\
\bottomrule
\end{tabular}
  \caption{The nRMSEs FNO with \cape in terms of each viscosity of 2D NS equations. Visualization is given in \autoref{fig:Generalization}}
  \label{tab:pde-param-2DNS}
\end{table}

\subsection{Curriculum Strategy Study}

In \autoref{tab:Ablation_full} we provide a full ablation study result of the curriculum strategy. 

\begin{table}[t]
\small
  \centering
  \begin{tabular}{llll}
  \toprule
  PDE & Model & Ablation &  nMSE \\
  \midrule
  \multirow{3}{*}{Advection} & \multirow{3}{*}{FNO} & curriculum strategy & ${\bf 0.04}$ \\

                  & & pure Autoregressive & $0.11$ \quad (${\bf + 0.07}$)\\
                  & & pure Teacher-Forcing & ${\bf 0.04}$ \quad ($\pm 0.00$)\\
                  \midrule
  \multirow{3}{*}{Advection} & \multirow{3}{*}{Unet} & curriculum strategy & ${\bf 0.11}$ \\
                  & & pure Autoregressive & $0.17$ \quad (${\bf + 0.06}$)\\
                  & & pure Teacher-Forcing & $0.12$ \quad ($+ 0.01$)\\
                  \midrule
 \multirow{3}{*}{Burgers} & \multirow{3}{*}{FNO} & curriculum strategy & ${\bf 0.13}$ \\
                  & & pure Autoregressive & $0.16$ \quad (${\bf + 0.03}$)\\
                  & & pure Teacher-Forcing & ${\bf 0.13}$ \quad ($+ 0.00$)\\
                  \midrule
 \multirow{3}{*}{Burgers} & \multirow{3}{*}{Unet} & curriculum strategy & ${\bf 0.45}$ \\
                  & & pure Autoregressive & $0.94$ \quad (${\bf + 0.49}$)\\
                  & & pure Teacher-Forcing & $0.84$ \quad ($+ 0.39$)\\
                  \midrule
  \multirow{3}{*}{2D NS} & \multirow{3}{*}{FNO} & curriculum strategy & ${\bf 8.0 \times 10^{-1}}$ \\
                  && pure Autoregressive & $1.3\times10^{+0}$  \quad ($+ 0.5$) \\
                  && pure Teacher-Forcing & $3.2\times10^{+0}$  \quad (${\bf + 2.4}$) \\                  
                  \midrule                  
  \multirow{3}{*}{2D NS} & \multirow{3}{*}{Unet} & curriculum strategy & ${\bf 7.0\times10^{-1}}$ \\
                  && pure Autoregressive & $1.0\times10^{+0}$  \quad (${\bf + 0.3}$) \\
                  && pure Teacher-Forcing & $1.0\times10^{+0}$  \quad (${\bf + 0.3}$) \\                  
  \bottomrule
  \end{tabular}
  \caption{Ablation study for the Advection, Burgers and 2D CFD equations with FNO as  \base model. }
   %Here AR is autoregressive and TR is teacher-forcing.}
  \label{tab:Ablation_full}
\end{table}

\section{Detailed Description of the Curriculum Strategy}

\begin{figure}[t!]
  \centering
  \includegraphics[width=0.5\textwidth]{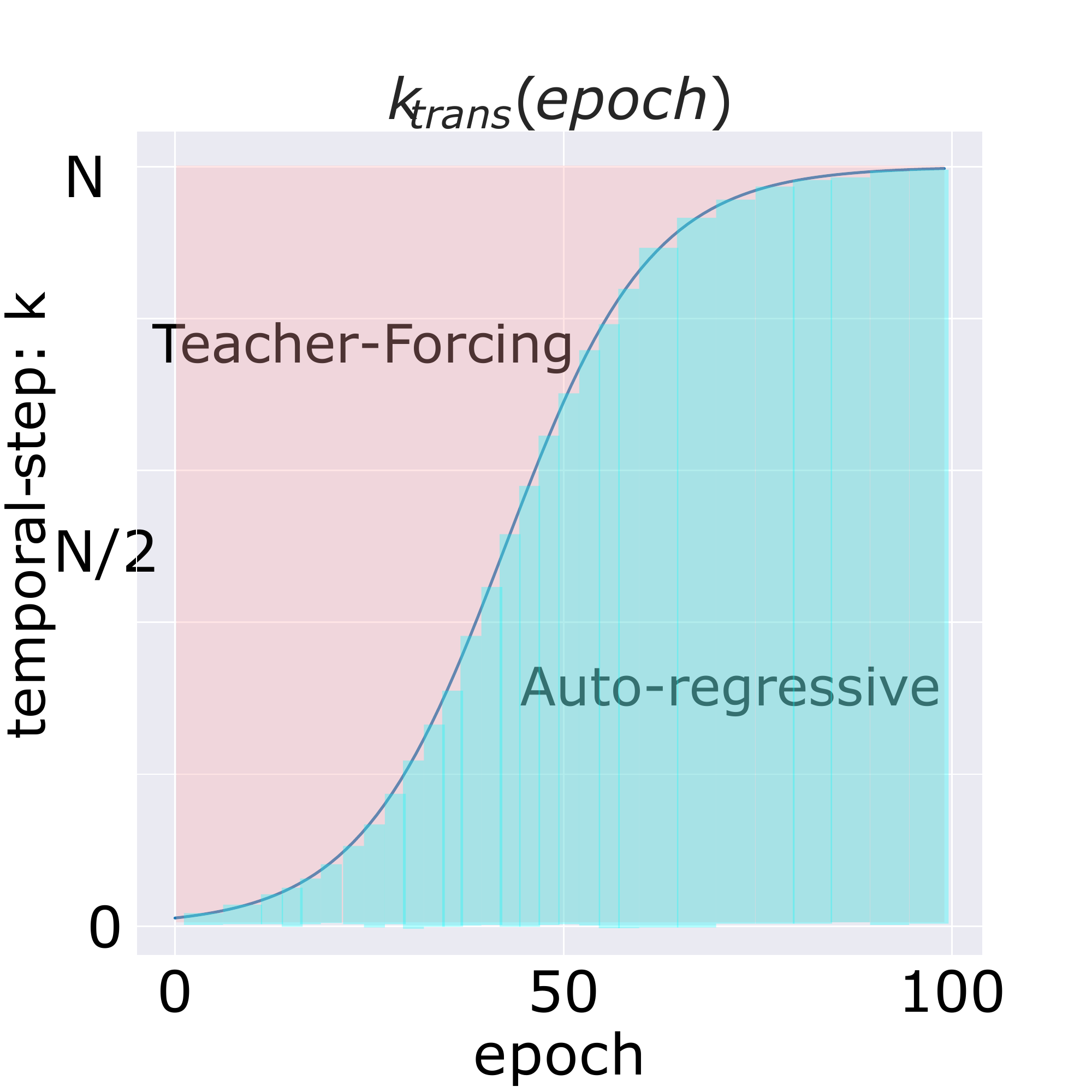}
  \caption{A plot of an instance of the function defined in \autoref{eq:CRC} where we set: $\Delta = 0.25$.}
  \label{fig:t-trans}
\end{figure}

\autoref{fig:t-trans} plots the profile of \autoref{eq:CRC} in terms of the epoch number where the maximum epoch number is assumed 100. We also provided the detailed algorithm of our curriculum strategy in terms of epochs and temporal steps in Algorithm \autoref{alg:CRC}. 
In our all the calculation with curriculum strategy, we set: $\Delta = 0.2$. 

%\textcolor{red}{
Concerning the form of $k_{\rm trans}$ in \autoref{eq:CRC}, it has many variations, such as a simple linear growth. We empirically decided the expression of it as given in \autoref{eq:CRC} which allows to rather gradual start and end of the transition process thanks to the form of hyperbolic tangent function.
%}

\begin{algorithm}[t]
  \caption{Algorithm of the curriculum training strategy}
  \label{alg:CRC}
  \textbf{Input} model parameters $\btheta$, training epoch number $n$, total training epoch number $M$, Training samples: $\{ \bu^{i} \}_{i=0,\cdots,N}$, temporal index $k$, final time step of the training sample $N$, $\epsilon$ is the random noise.  
  \begin{algorithmic}[1]
  \FOR{n $= 0$ {\bf to} $M$}
    \STATE Calculate $k_{\rm trans}$ following \autoref{eq:CRC},    
    \FOR{k $= 0$ {\bf to} $N - 1$}
      \IF{$k \leq k_{\rm trans}$}
        \STATE $\tilde{\bu}^{k+1} = \mathrm{NN}(\tilde{\bu}^{k} + \epsilon; \btheta)$ %\quad  \leftarrow \quad (\mathrm{auto \ regressive})
      \ELSE
        \STATE $\tilde{\bu}^{k+1} = \mathrm{NN}(\bu^{k} + \epsilon; \btheta)$ \quad  %\leftarrow \quad (\mathrm{teacher \ forcing})
      \ENDIF
    \STATE $\mathbf{L}^{k} \leftarrow \mathrm{MSE}(\tilde{\bu}^{k+1}, \bu^{k+1})$
    \ENDFOR
    \STATE $\btheta \leftarrow$ Optimizer $(\sum_{k=1}^{N} \mathbf{L}^{k})$
  \ENDFOR    
  \end{algorithmic}
\end{algorithm}

%\section{\label{sec:ablation_CAPE_structure}Ablation Study for CAPE module structure}
\section{\label{sec:ablation_CAPE_structure}Study for CAPE module structure}

\subsection{Ablation Study}

In this section, we provided results of ablation study for our CAPE module's internal structure to provide an insight of the inductive bias of CAPE. 
In this study, we performed training without (1) spectral-convolution, (2) $1 \times 1$-convolution, and (3) depthwise-convolution. The results were provided in \autoref{tab:ablation_CAPE_structure} that indicates that  all the 3-convolution layers and LayerNormalization play important roles on the error, but the spectral-convolution has the strongest impact. However, it also shows that the important factor depends on PDEs because of the difference of PDE natures (e.g., advection, diffusion, or non-linear system equations, and so on). It also indicates that our selection always shows a better result, though not always the best. 

\begin{table}[h]
  \centering
  \begin{tabular}{lcll}
  \toprule
   PDE & model & Ablation & nRMSE  \\
    \midrule
    1D Advection & FNO & \base & 0.04 \\
    &  & w/t Spectral Convolution  & 0.06 \ \textbf{(+0.02)} \\
    &  & w/t $1 \times 1$ Convolution & 0.05 \ (+0.01) \\
    &  & w/t Depthwise Convolution & 0.05 \ (+0.01) \\    
    &  & w/t LN & 0.03 \ (-0.01) \\        
    \midrule
    1D Burgers & FNO & \base & 0.13 \\
    &  & w/t Spectral Convolution  & 0.13 \ ($\pm 0.00$) \\
    &  & w/t $1 \times 1$ Convolution & 0.12 \ (-0.01) \\
    &  & w/t Depthwise Convolution & 0.13 \ ($\pm 0.00$) \\    
    &  & w/t LN & 0.09 \ (-0.04) \\        
    \midrule
    2D NS & FNO & \base & 0.80 \\
    &  & w/t Spectral Convolution  & 0.91 \ (+0.11) \\
    &  & w/t $1 \times 1$ Convolution & 0.80 \ ($\pm 0.00$) \\
    &  & w/t Depthwise Convolution & 0.76 \ (- 0.04) \\    
    &  & w/t LN & 1.24 \ \textbf{(+0.45)} \\        
\bottomrule
\end{tabular}
\caption{Ablation study of CAPE internal Structure.}
\label{tab:ablation_CAPE_structure}
\end{table}

%\color{red}
\subsection{Study on Other Possibility of CAPE Structure}

Inspired by recent work \cite{tu2022maxvit}, we also tried a sequential manner of the convolution layers in the CAPE module.  as described in \autoref{eq:cape_conv}. 

\begin{align}
    \by^k_{\alpha}(x) &= h_{1\times 1, d \rightarrow c \times \ell} \left(\sigma\left(h_{1 \times 1, c \rightarrow d}(\bu^{k}) + \bv_\alpha^k \right) \right) (x), 
    \\
    \by^k(x) &= \by^k_{\alpha_1}(\by^k_{\alpha_2}(\by^k_{\alpha_3}(x))),
    \label{eq:cape_conv_v2}
\end{align}
where $x$ is the output of the previous layer or input, and $\alpha_k$ is either spectral convolution, $1 \times 1$ convolution, or depthwise convolution. The result on 2D NS equations is provided in \autoref{tab:CAPE_structure_v2}. It shows that in this case the order (depthwise conv., $1 \times 1$ conv., spectral conv.) is the best choice. However, it also shows that the best error is still larger than the vanilla CAPE module if the model weight parameter number is the same. 

\begin{table}[h]
  \centering
  \begin{tabular}{lclll}
  \toprule
   PDE & model & \# model parameters & module order & nRMSE  \\
    \midrule
    2D NS & FNO & 1.16M & ($1 \times 1$,D,S) & 0.78 \\
    &  &  &  ($1 \times 1$,S,D)  & 0.77  \\
    &  &&  (D,$1 \times 1$,S) & {\bf 0.64}  \\
    &  &&  (D,S,$1 \times 1$) & 0.71 \\    
    &  &&  (S,$1 \times 1$,D) & 0.73 \\        
    &  &&  (S,D,$1 \times 1$) & 0.82 \\        
    \midrule
    2D NS & FNO & 0.83M &  (D,$1 \times 1$,s) & 0.79  \\   
\bottomrule
\end{tabular}
\caption{Sequential type CAPE internal Structure. (S, $1 \times 1$, D) mean (spectral conv., $1 \times 1$ conv., depthwise conv.).}
\label{tab:CAPE_structure_v2}
\end{table}

\section{Larger Time-step Experiments\label{sec:larger-timestep}}

%In our experiments, the time-step size of the data might be somewhat small (1D: 40, 2D: 20). To clarify the ability of \cape in larger time-step cases, we performed additional experiments with 100-time step ($\Delta t = 0.02$). The results are provided in \autoref{tab:larger_timesteps} which indicates that \cape still shows an ability to understand the PDE parameters, though the nRMSE is a little larger than the smaller time step number case. 

%\textcolor{red}{
In our experiments, the time-step size of the data might be somewhat small (1D: 40, 2D: 20). This small number of time step, however, might be attributed to the effectiveness of \cape and the curriculum strategy because the small number of time steps may reduce the accumulation error in time due to the autoregressive nature of the model’s prediction, which can result in training instability and a lack of generalizability. To clarify the robustness of \cape and the curriculum strategy in the light of the time-step number, we performed additional experiments with 100-time steps ($\Delta t = 0.02$). The results are provided in \autoref{tab:larger_timesteps} which evinced that both FNO and FNO w.t. \cape were trained effectively, without any instability or severe error accumulation, though the errors themselves are moderately increased relative to the case with 40 steps.
%}

\begin{table}[h]
  \centering
  \begin{tabular}{lll}
  \toprule
   PDE & model & nRMSE  \\
    \midrule
    1D Advection & FNO & 0.90 \\
    &  FNO with \cape & 0.18  \\
\bottomrule
\end{tabular}
\caption{nRMSE of the case with larger time-steps on 1D Advection equation using FNO. In the experiments, we trained the FNO/FNO with \cape with a larger time step number: 100 steps. }
\label{tab:larger_timesteps}
\end{table}
%\color{black}
%\end{comment}

\section{Visualization of Attention Weights\label{sec:vis-attn}}

\begin{figure}[t!]
  \centering
  \includegraphics[width=\textwidth]{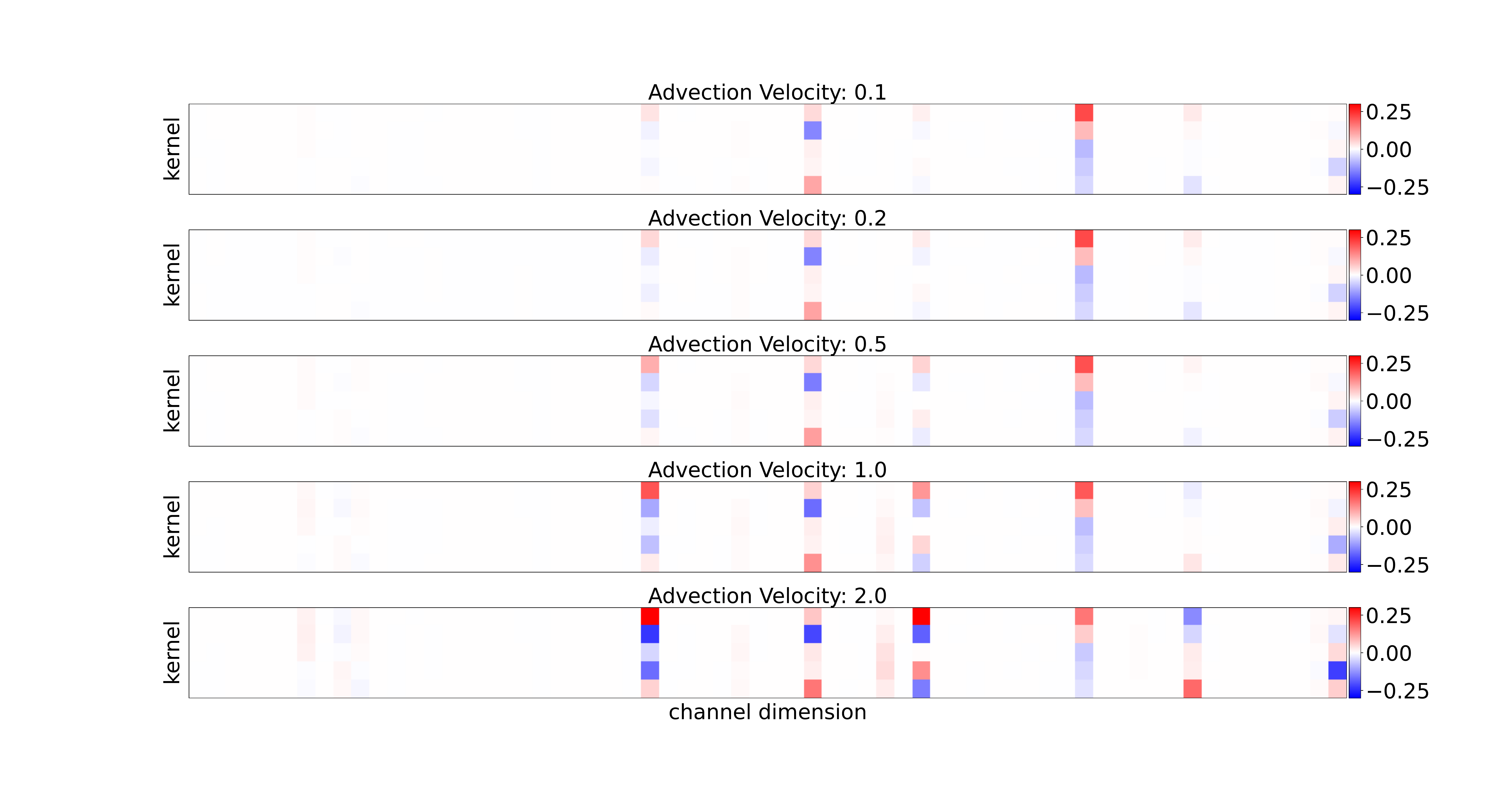}
  \caption{Plot of attention weights in the case of 1D Advection equation.}
  \label{fig:vis-adv-attn}
\end{figure}

\begin{figure}[t!]
  \centering
  \includegraphics[width=\textwidth]{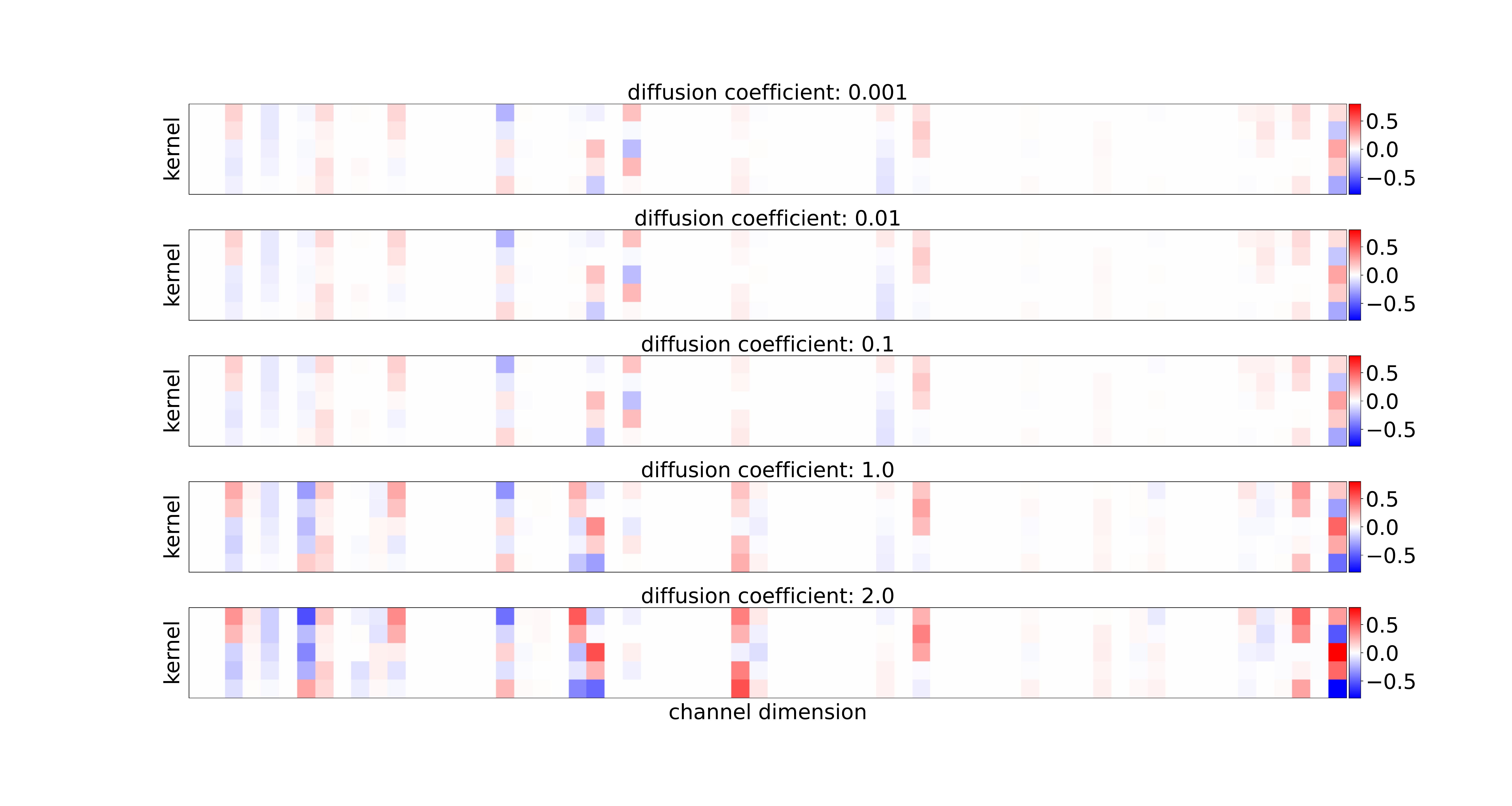}  
  \caption{Plot of attention weights in the case of 1D Burgers equation.}
  \label{fig:vis-bgs-attn}
\end{figure}

\autoref{fig:vis-adv-attn} and \autoref{fig:vis-bgs-attn} are the plots of the Convolution kernel after being multiplied by the channel attention. For simplicity, we only consider the one-dimensional cases and the depthwise convolution kernel that is more interpretable than the other convolution kernels. \autoref{fig:vis-adv-attn} is the case of 1D Advection equation. It shows that the kernel is very sparse and the channel attention chooses different kernel as the advection velocity increases, resulting in adjusting appropriate information transfer corresponding with the input advection velocity. \autoref{fig:vis-bgs-attn} is the case of 1D Burgers equation. It shows that the kernel becomes nearly zero by channel attention when the diffusion coefficient is small, indicating a small diffusion. On the other hand, the number of activated kernels increases with the diffusion coefficient, indicating that the operation becomes closer to the averaging, corresponding with the function of the diffusion. 


\begin{comment}
{\color{blue}
\section{Connection to the Implicit Schema}

The implicit method from the numerical literature solves for $u^{k+1}$ according to 
$$  u^{k+1} = u^k + \Delta t F(u^{k+1}; \lambda) $$
% \equiv \tilde{F}(u^k, u^{k+1}; \lambda) $$
with CAPE we provide a surrogate prediction for $u^{k+1}$ in the right hand-side term. In this way, we do not need to solve for $u^{k+1}$ iteratively.
}

% \begin{equation}
%   \bu^{k+1} = \bu^k + \Delta t F(\bu^{k+1}; \blambda) \equiv \tilde{F}(\bu^k, \bu^{k+1}; \blambda)
%   \label{eq:pde_implicit_annex}
%   .
% \end{equation}
\end{comment}

%%%%%%%%%%%%%%%%%%%%%%%%%%%%%%%%%%%%%%%%%%%%%%%%%%%%%%%%%%%%%%%%%%%%%%%%%%%%%%%
%%%%%%%%%%%%%%%%%%%%%%%%%%%%%%%%%%%%%%%%%%%%%%%%%%%%%%%%%%%%%%%%%%%%%%%%%%%%%%%

\end{document}